\documentclass[fleqn,10pt]{wlscirep}
\usepackage[utf8]{inputenc}
\usepackage[T1]{fontenc}

\usepackage{cite}
\usepackage{amsmath,amssymb,amsfonts}
\usepackage{algorithmic}
\usepackage{graphicx}
\usepackage{bm}
\usepackage{float}
\usepackage{textcomp}
\usepackage{xcolor}
\usepackage{subfigure}
\usepackage{url}
\usepackage{algorithm}
\usepackage{algorithmic}
\usepackage{makecell}
\usepackage{color}
\usepackage{verbatim}
\usepackage{hyperref}
\usepackage{wrapfig}
\usepackage{multicol}
\usepackage{mathptmx}

\title{\LARGE FedPEAT: Convergence of 6G Enabled Federated Learning, Parameter-Efficient Fine Tuning, and Emulator Assisted Tuning for AI Foundation Models}

\author[1,+]{Terence Jie Chua}
\author[1,+]{Wenhan Yu}
\author[1]{Yang Li}
\author[2,*]{Jun Zhao}
\affil[1]{Nanyang Technological University, Graduate College, Singapore, 637335, Singapore}
\affil[2]{Nanyang Technological University, School of Computer Science and Engineering, Singapore, 639798, Singapore}

\affil[*]{junzhao@ntu.edu.sg}

\affil[+]{These authors contributed equally to this work.}

\begin{abstract}
The advent of foundation models like \mbox{GPT-3} and BERT has revolutionized artificial intelligence, providing unparalleled
capabilities across various applications and the potential to transform industries from healthcare to entertainment. Deploying
and fine-tuning these models pose unique challenges, making it imperative to address issues like model ownership, collaborative
training, and computation and communication limitations for realizing their full potential. We generalize the offsite tuning
approach to Emulator-Assisted Tuning (\mbox{\texttt{EAT}}) and combine it with Parameter-Efficient Fine-Tuning (PEFT) to create Parameter-
Efficient Emulator-Assisted Tuning (\mbox{\texttt{PEAT}}), expanding its use into 6G-enabled Federated Learning (FL) as Federated Parameter-
Efficient Emulator-Assisted Tuning (\mbox{\texttt{FedPEAT}}). The \mbox{\texttt{FedPEAT}} framework proposes a solution using adapters, emulators, and
PEFT techniques for federated model fine-tuning. This approach enhances model privacy and streamlines downstream fine-tuning. Our approach, adaptable to diverse
neural network architectures,
 incorporates an adaptive control mechanism utilizing the novel Single-Agent Action Branching
Proximal Policy Optimization (SABPPO) algorithm. The proposed SABPPO is tailored for high-dimensional action spaces, featuring short training delays, essential for scalable \mbox{\texttt{FedPEAT}} involving a large number of users
and variables to optimize. Our experimental results demonstrate the practicality and efficacy of our proposed
framework and algorithm in addressing the complex challenges associated with large foundation model fine-tuning.

\end{abstract}
\begin{document}

\flushbottom
\maketitle
\thispagestyle{empty}

\noindent In the vibrant landscape of artificial intelligence (AI), colossal foundation models like \mbox{GPT-3}~\cite{brown2020language}, CLIP~\cite{radford2019language} and BERT~\cite{devlin2018bert} have revolutionized AI, venturing beyond traditional machine learning approaches. These models, trained on massive datasets, possess the uncanny ability to generate images, texts, and audio with unparalleled accuracy. With sizes reaching billions of parameters, they capture intricate linguistic nuances and showcase human-level proficiency across diverse applications. Large foundation models have garnered attention for their capacity to adapt to new tasks and domains through a transfer learning approach called fine-tuning~\cite{wei2021finetuned,muennighoff2022crosslingual}. Leveraging these models offers an advantage in terms of time and resource savings as compared to training models from the ground up, especially for large models like GPT3 with 175B+ parameters. The dawn of 6G technologies, boasting broad bandwidths (1 THz to 3THz) and unprecedented data communication speeds~\cite{letaief2019roadmap}, potentially reaching terabits per second, opens avenues for federated fine-tuning of these expansive models.

\begin{figure}[htbp] 
\centering
\setlength{\abovecaptionskip}{-0.0cm}
\includegraphics[width=0.5\linewidth]{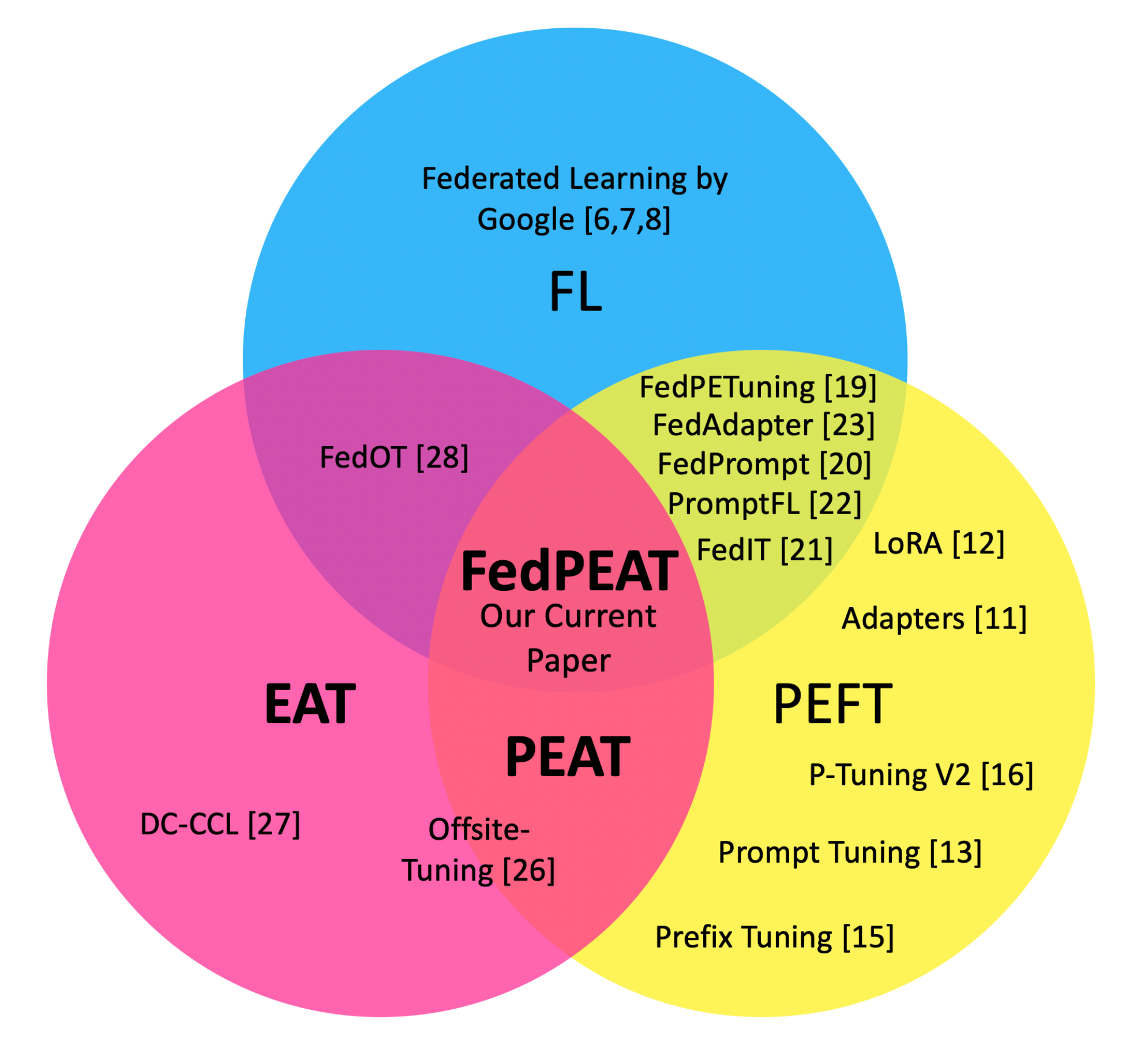}
\caption{Intersection of Federated learning (FL), Parameter-Efficient Fine-Tuning (PEFT), and Emulator-Assisted Tuning (\mbox{\texttt{EAT}}). Here we illustrate the intersection of FL, PEFT, and (\mbox{\texttt{EAT}}). The main contribution of our current paper is to introduce Federated Parameter-Efficient Emulator-Assisted Tuning (\mbox{\texttt{FedPEAT}}), as a convergence of \mbox{\texttt{EAT}}, PEFT, and FL, while \mbox{\texttt{EAT}} and Parameter-Efficient Emulator-Assisted Tuning (\mbox{\texttt{PEAT}}) are also terms coined by our paper.}
\vspace{-0.4cm}
\label{fig:Venn}
\end{figure}

One of the significant challenges associated with fine-tuning large language models lies in the distribution of data. Many real-world applications necessitate the utilization of data that resides on user devices, such as smartphones, laptops, and mobile edge devices, rather than on centralized servers. The need for decentralization hinders foundation model fine-tuning. Federated learning (FL) has emerged as promising solutions to these issues. Federated learning is a decentralized machine learning approach that enables privacy-preserving model training without the need to centralize data~\cite{mcmahan2017communication,konevcny2016federated, bonawitz2019towards}. Instead of sending raw data to a central server, federated learning trains models directly on the user's device. These models are then aggregated to create a global model, preserving data privacy while achieving the desired performance. 

However, fine-tuning large language models is computationally intensive. Training a model with hundreds of millions or billions of parameters demands substantial computational resources, often beyond the reach of individual users or small organizations. This computational bottleneck can limit the widespread adoption of these models and impede their deployment in resource-constrained environments. Moreover, fine-tuning on local devices, such as smartphones or edge devices, is often not feasible due to their limited computational capabilities. Distributing the model fine-tuning process across devices while ensuring data privacy and model performance adds another layer of complexity. In response to these challenges, various methods have been explored to make fine-tuning of pre-trained models more efficient. Efforts in model tuning have extended to the realm of adapters~\cite{rebuffi2017learning,he2021towards, houlsby2019parameter}, which encode task-specific representations within intermediate layers while preserving pre-training knowledge. Different Parameter-Efficient Fine Tuning (PEFT) techniques have been proposed, encompassing approaches such as Low Rank Adapters (LoRA)~\cite{hu2021lora}, prompt tuning~\cite{qin2021learning,lester2021power}, prefix-tuning~\cite{li2021prefix}, adapters~\cite{houlsby2019parameter}, P-tuning V2~\cite{liu2021p}, tuning embedding layer inputs~\cite{an2022input}, tuning hidden states~\cite{liu2022few}, and more. These methods aim to update or add only a limited number of model parameters, reducing resource requirements and allowing for the sharing of parameters from the pre-trained model. Several authors of the works~\cite{zhang2023fedpetuning, zhao2023fedprompt, zhang2023towards, guo2023promptfl, cai2022fedadapter} noticed the prowess of PEFT techniques and proposed Federated-PEFT approaches.

However, large language models are often owned by research institutions or companies that bear the responsibility of maintaining and updating them. These model owners typically cannot directly share the entire model with external devices due to various reasons, including privacy concerns, intellectual property rights, and the potential for misuse. The lack of easy sharing mechanisms hampers the democratization of large language models and their use in applications that require continuous updates and fine-tuning. As a result, there is a need to develop mechanisms that allow model owners to collaborate with external parties or distribute portions of models securely. Federated fine-tuning for downstream tasks of local devices often necessitates knowledge of the entire model's weights, potentially raising privacy concerns. Furthermore, the process of fine-tuning and deploying foundation models can pose significant resource challenges due to their substantial parameter sizes~\cite{smith2022using,xiao2023smoothquant}. Xiao~\textit{et~al.}~\cite{xiao2023offsite} proposed an approach to fine-tune large foundation models using the combination of an emulator, which is a compressed version of a subset of the original large foundation model, and an adapter, which are the trainable weight to be shared. Nevertheless, these authors do not consider the federated and collaborative tuning between devices. Ding~\textit{et~al.}~\cite{ding2023dc} introduced an approach that involves model compression and an emulator-adapter-like strategy for collaborative tuning of large vision models in a device-server setting. Kuang~\textit{et~al.}~\cite{kuang2023federatedscope} proposed FedOT, which is federated version of offsite-tuning. Although Kuang~\textit{et~al.}~\cite{kuang2023federatedscope} briefly touch upon an architecture similar to our proposed Federated Parameter-Efficient Emulator-Assisted Tuning (\mbox{\texttt{FedPEAT}}), they do not provide detailed discussions. 

\iffalse
In light of these, we propose the \mbox{\texttt{FedPEAT}} framework which generalizes the emulator-adapter approach to all architectural configurations for the federated learning context, and proposed an adaptive control mechanism to support the adoption of \mbox{\texttt{FedPEAT}}.
\fi

\begin{figure}[t] 
\centering
\setlength{\abovecaptionskip}{0.5cm}
\includegraphics[width=0.85\linewidth]{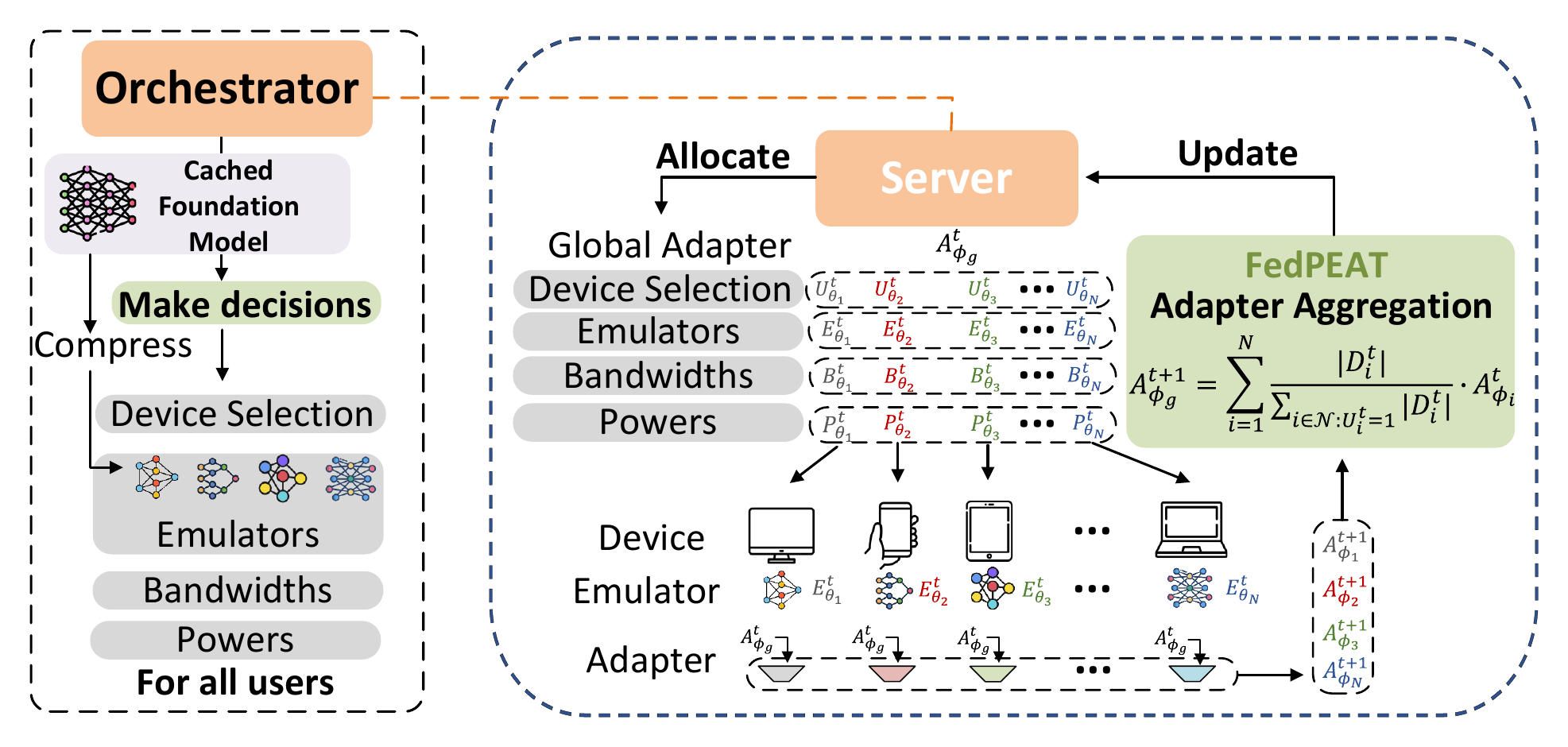}
\caption{ \mbox{\texttt{FedPEAT}} with Adaptive control overview. This figure shows how the Adaptive control orchestrator makes decisions on important parameters, such as device selection, emulator compression parameter, transmission bandwidth and power to facilitate the \mbox{\texttt{FedPEAT}} process.}
\vspace{-0.3cm}
\label{fig:overview}
\end{figure}

\section*{Overview of \mbox{\texttt{FedPEAT}} with adaptive control}

\subsection*{Proposed \mbox{\texttt{EAT}} structure}
In addressing the pressing issues of model and data privacy and ownership, as well as the imperative need for memory and computation-efficient downstream model fine-tuning, we propose a novel Emulator-Assisted Tuning (\mbox{\texttt{EAT}}) structure which generalizes the offsite tuning approach introduced by Xiao et al.~\cite{xiao2023offsite} to encompass all possible combinations of adapter and emulator configurations for large foundation model fine-tuning. Our proposed \mbox{\texttt{EAT}} structure offers the flexibility to adapt the adapter and emulator to the specific requirements of a given application. The adapter and emulator can take any form, whether encompassing layers within a transformer architecture, multi-layer perceptron, or any other neural network structure. The emulator, can have variable number of neural network layers, variable number of nodes per layer, and even variable arrangements of transformer attention units. This adaptability ensures that the model can be fine-tuned efficiently across a wide spectrum of tasks, from simple to complex.

\subsection*{Expansion to \mbox{\texttt{PEAT} architecture}}
In the field of model tuning, various Parameter-Efficient Fine Tuning (PEFT) methods such as Low-rank Adapters (LoRA)~\cite{hu2021lora}, prompt tuning~\cite{qin2021learning}, and adapters~\cite{rebuffi2017learning,he2021towards, houlsby2019parameter} have been explored to make fine-tuning of pre-trained models more efficient. We combine \mbox{\texttt{EAT}} and Parameter-Efficient Fine-Tuning (PEFT) to present Parameter-Efficient Emulator-Assisted Tuning (\mbox{\texttt{PEAT}}).

\subsection*{\mbox{\texttt{FedPEAT}} framework}
We extend the use of the \mbox{\texttt{PEAT}} into the domain of Federated Learning (FL) and introduce a novel framework, Federated Parameter-Efficient Emulator-Assisted Tuning (\mbox{\texttt{FedPEAT}}). This unique integration not only addresses model and data privacy concerns by eliminating the need for the model owner to transmit the entire model to the client and the client to send their local data to the model owner but also substantially improves the memory and computational efficiency of collaborative, downstream federated model fine-tuning. We illustrate the intersection our proposed \mbox{\texttt{EAT}}, \mbox{\texttt{PEAT}}, and \mbox{\texttt{FedPEAT}} in Figure~\ref{fig:Venn}.

% a novel \mbox{\texttt{FedPEAT}} framework that combines the use of adapters and emulators (shown in Fig.~\ref{fig:overview}). Adapters, comprising a few layers of trainable neural network parameters, enable the customization of a model to a specific task, making it a versatile tool for federated model fine-tuning. Emulators, on the other hand, constitute parts of the foundation model neural network with fixed parameters, offering a compressed version of the original model. This innovative combination not only mitigates model privacy concerns, as the model owner does not have to transmit the entire model to client's local edge devices, but also significantly enhances the memory and computation efficiency of downstream model fine-tuning.

% \textbf{\mbox{\texttt{PEAT}} architecture. }Furthermore, our proposed \mbox{\texttt{FedPEAT}} framework offers the flexibility to adapt the adapter and emulator to the specific requirements of a given application. The adapter and emulator can take any form, whether encompassing layers within a transformer architecture, multi-layer perceptron, or any other neural network structure. The emulator, can have variable number of neural network layers, variable number of nodes per layer, and even variable arrangements of transformer attention units. This adaptability ensures that the model can be fine-tuned efficiently across a wide spectrum of tasks, from simple to complex.
\subsection*{\mbox{\texttt{FedPEAT}} adaptive control mechanism}
To optimize and streamline this adaptive combination of adapters and emulators, we propose coupling them with an adaptive control mechanism. This mechanism employs a deep reinforcement learning orchestrator to control critical hyper-parameters, such as emulator model compression ratios, adapter parameter-efficient fine-tuning parameters, and even device selection for participation in collaborative federated learning during each iteration (shown in Figure~\ref{fig:overview}). This integration facilitates the efficient orchestration of resources, ensuring that the fine-tuning process remains memory, computation, and communication-efficient. This orchestration ensures that participating devices possess the necessary computational resources to carry out fine-tuning effectively. This contribution is essential in guaranteeing the successful application of our model adaptation and fine-tuning technique in real-world, resource-constrained environments.

\subsection*{Server-Device collaborative tuning}
The \mbox{\texttt{FedPEAT}} framework is applicable to collaborative FL of various contribution nature. We note two distinct types of contribution cases. The first case involves FL where all data resides on mobile edge devices (i.e., clients), with no central server involvement. In this scenario, model tuning is entirely performed on the client, while the server's role is restricted to aggregating adapter module parameters. The second case entails federated learning where data is distributed across both client devices and a central server. Fine-tuning occurs on both client devices and the server, presenting a more complex but realistic setting that highlights the adaptability and versatility of our proposed framework. In our experiments, we consider the special case of \mbox{\texttt{FedPEAT}} framework in which the server possesses data and partakes in the collaborative federated foundation model fine-tuning process instead of acting purely as an aggregator. Through these experiments, we aim to demonstrate the practical applicability and efficacy of our approach.

\section*{Parameter-Efficient Emulator-Assisted Tuning (\mbox{\texttt{PEAT}}) Sub-units}
\iffalse
In this section, we explore the intricacies of the \mbox{\texttt{PEAT}} sub-components, specifically the emulator, adapter modules, and the Parameter-Efficient Fine-Tuning (PEFT) adaptation.
\fi

\subsection*{Emulator}
The emulator represents a collection of neural network weights meticulously designed to mimic the behavior of the original foundation model. Through the compression of extensive neural network knowledge into a more compact architecture, emulators aim to deliver performance that closely rivals their larger counterparts while dramatically reducing computational and storage requirements. The decision to share emulators with client devices, rather than the original foundation model, serves a dual purpose: firstly, it safeguards the proprietary nature of model ownership by obviating the need to divulge the complete model to local devices; secondly, it empowers local devices to store and undertake model fine-tuning using a significantly smaller-sized emulator. In essence, an emulator serves as a streamlined and resource-efficient rendition of a more extensive model, crafted through techniques such as pruning~\cite{han2015deep}, layer drop~\cite{sajjad2023effect}, or knowledge distillation~\cite{hinton2015distilling}. Importantly, our approach employs emulators with fixed-parameter values, without fine-tuning, to encapsulate the bulk of knowledge and information derived from pre-trained foundation models. 

% In the ever-evolving landscape of deep learning, the increasing size of state-of-the-art models has presented both opportunities and challenges. While large models exhibit impressive performance across a number of tasks, their computational requirements can be prohibitive for deployment in resource-constrained environments, especially for model devices~\cite{llmondevice}. To address this disparity, the concept of an ``emulator'' has emerged. An emulator is a compact and efficient version of a more extensive model, obtained through techniques such as pruning, layer drop or knowledge distillation~\cite{modelcompression}. By condensing the knowledge of a vast neural network into a smaller architecture, emulators aim to offer a near-par performance of their larger counterparts while significantly reducing the computational and storage overhead. Such an approach bridges the gap between high-performing models and real-world, on-the-ground applicability, ushering in a new era of efficient deep learning.

\subsection*{Adapter}
% Transfer learning has become a fundamental technique in deep learning, offering a pathway to leverage the knowledge of pre-trained models for new tasks.~\cite{transferlearning}. However, fine-tuning an entire pre-trained model can be computationally expensive and may require substantial amounts of data. This is where the notion of ``adapters'' comes into play.

Adapters are modular additions to pre-existing foundation models like large language model (LLM), designed to facilitate task-specific adaptations with minimal modifications to the original model~\cite{xiao2023offsite}. Essentially, adapters are a smaller set of neural network weights with tunable parameters so as to encode information at the user device for downstream task fine-tuning. The smaller adapter size serves two main purpose. Firstly, the adapter is designed to be a plug which can be conveniently placed at the end of the original foundation model at the server and also a plug at the end of the emulator on the local devices. Secondly, the smaller adapter size reduces adapter transmission costs. By only tuning the parameters of these added layers, one can harness the generalized capabilities of large models while efficiently tailoring them for specific tasks. 

\subsection*{PEFT integration}
PEFT methods like LoRA~\cite{hu2021lora} and Adapter~\cite{houlsby2019parameter} can significantly reduce model size, consequently save memory, while achieving comparable model performance to a model which do not use PEFT approaches~\cite{xiao2023offsite}. The integration of PEFT methods is seamless and can be directly applied on the adapter module in each federated learning iteration.
% The adapter methodology underscores the power of modular and scalable deep learning, enabling efficient multi-task learning and swift adaptation.

\section*{Federated Parameter-Efficient Emulator-Assisted Tuning (\mbox{\texttt{FedPEAT}}) Framework}
\iffalse
In this section, we will discuss our proposed innovative framework, known as the Federated Parameter-Efficient Emulator-Assisted Tuning (\mbox{\texttt{FedPEAT}}). We offer a thorough examination of the synergistic interaction between emulators and adapters, illuminating their seamless integration with parameter-efficient tuning techniques and their expansion into the domain of federated learning.
\fi
\begin{figure*}[t] 
\centering
\setlength{\abovecaptionskip}{-0.1cm}
\includegraphics[width=1\linewidth]{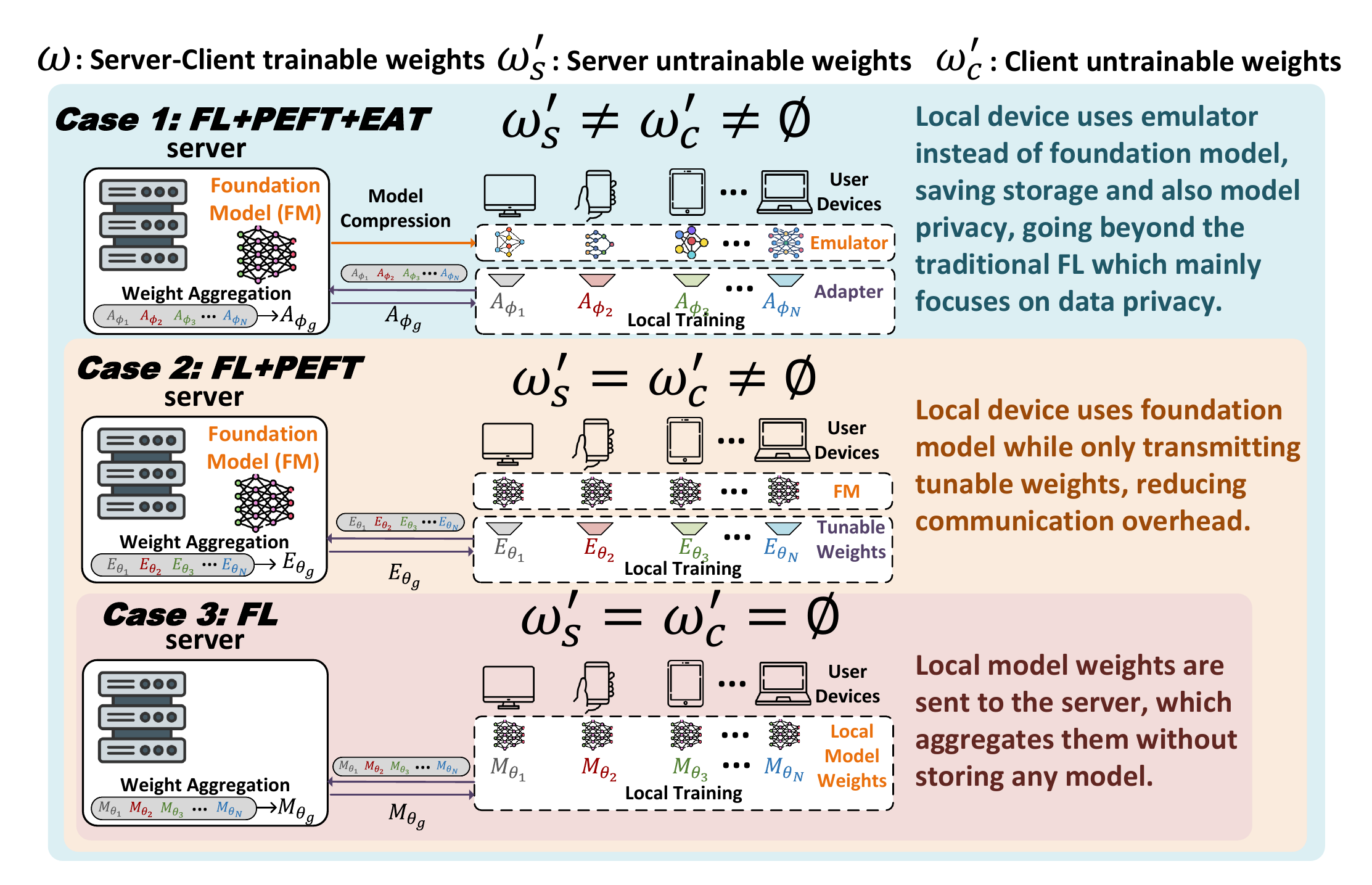}
\caption{Emulator-Assisted Tuning generalized to three cases. Figure illustrates how the neural network structures at the server and local devices differ in each case. Case 1 represents our proposed \mbox{\texttt{FedPEAT}} framework. Case 2 represents the integration of Federated Learning and PEFT. Case 3 represents a traditional Federated Learning scenario.}\vspace{-10pt}
\label{fig:systemmodel}
\end{figure*}

\subsection*{Emulators and Adapters}
The server houses a foundation model $M_{\theta_g}$, while each user device (UEs) labeled by index $n$ receives and holds:
\begin{align}
    \begin{cases}
        \text{An adapter $A_{\phi_n}$ specifically tuned for the downstream task.} \\
        \text{An emulator $E_{\theta_n}$, which is a tailored version of the} \\ \text{~~~~~~~~~~~~~~~~~~~~~~foundational model, represented by} \\ \text{~~~~~~~~~~~~~~~~~~~~~~$E_{\theta_n}=f(M_{\theta_g} - A_{\phi_n})$.} \nonumber
    \end{cases}
\end{align}

The adapter, denoted as $A$, aligns with the definition put forth by~\cite{xiao2023offsite}, comprising sets of layers embedded within the foundation model's architecture. These layers feature tunable parameters, specifically designed to facilitate model fine-tuning by encoding new information from downstream tasks. On the contrary, the emulators denoted by $E$ encapsulate a version of the original foundation model that may have undergone modifications. The adaptation of the emulator occurs after the removal of the adapter layers and serves as a guiding framework for tuning the adapter parameters. The parameter value of the emulators are fixed and aims to emulate the large foundation models. The transformation function $f()$, in this context, refers to model compression algorithms such as layer dropping~\cite{zhang2020accelerating}, model pruning~\cite{han2015deep}. Let $\omega$ represent the weights that are collaboratively trainable on both the server and device. $\omega'_s$ refers to the untrainable weights specific to the server, excluding $\omega$. $\omega'_c$ denotes the untrainable weights specific to a device, distinct from $\omega$.

Given this, we can generalize emulator-assisted tuning to three cases:
\begin{itemize}
    \item \textbf{Case 1: }
    $\omega'_s \neq \omega'_c \neq \varnothing:$ This is our proposed, more generalized framework, in which we permit various user devices (UEs) to employ distinct emulators, denoted as $E_{\theta_n}$. These emulators correspond to the untrainable weights on a device, $\omega'_c$, which are maintained at fixed values. Similarly, the subset of the model with untrainable parameters $M'_{\theta}$ corresponds to $\omega'_s$. This flexibility is particularly important since UEs frequently operate with constrained storage and computational resources. Therefore, emphasizing the efficient decompression and adaptiveness of the foundation model becomes essential.
    \item \textbf{Case 2: } 
    $\omega'_s = \omega'_c \neq \varnothing:$ This scenario is a subset of Case 1. Here, the emulator designated for UE $n$ aligns with the static parameters of the overarching foundation model (i.e., $\omega'_s=\omega'_c$). In this setup, we synergize Federated Learning (FL) training with Parameter-Efficient Fine-Tuning (PEFT) techniques, reflecting strategies showcased in previous research such as ~\cite{FedPET, FedGPT, zhao2023fedprompt, guo2023promptfl}.
    \item \textbf{Case 3: } $\omega'_s = \omega'_c = \varnothing:$ This is another specific instance within the purview of Case 1. In this scenario, all participants utilize the adjustable parameters of the global foundation model $\omega$. Essentially, no weights remain untrainable beyond the collaboratively trainable ones (i.e., $\omega'_s = \omega'_c = \varnothing$). This methodology closely mirrors the conventional federated learning (FL) paradigm, where individual model parameters are amalgamated to shape the global model.
\end{itemize}
The details of the cases are further illustrated in Figure~\ref{fig:systemmodel}.

\subsection*{Tuning Process}
The server model, denoted as $M_{\theta_g}$, can be decomposed into two primary components: the untrainable subset of weights of the foundation model $M'_{\theta}$, and the adapter $A_{\phi}$. After such decomposition, the server model is expressed as $M'_{\theta} \circ A_{\phi}$, with the symbol ``$\circ$" signifying the neural network connections between $M'_{\theta}$ and $A_{\phi}$. It is important to note that the arrangement of layers $M'_{\theta}$ and $A_{\phi}$ is flexible and can be configured in various orders. Emulator-assisted tuning (\mbox{\texttt{EAT}}) is an approach which generalizes all emulator-adapter based configurations, which include those proposed by~\cite{xiao2023offsite} and extends it to cases beyond those proposed by~\cite{xiao2023offsite} such as the ``Vertical'' splitting of the foundation model~\cite{ding2023dc}. Furthermore, the term ``\textit{offsite}'' in their work~\cite{xiao2023offsite} only considers a single device tuning and does not consider a multiple device collaborative training scenario. Our proposed \mbox{\texttt{EAT}} approach generalizes the emulator and adapter approach to collaborative tuning between multiple devices. Furthermore, Xiao~\textit{et~al.}~\cite{xiao2023offsite} do not consider a collaborative fine-tuning scenario where there are datasets that are stored at the server and that the server is able to partake in collaborative fine-tuning as proposed by~\cite{ding2023dc}. The emulator-to-be, represented as $M'_{\theta}$, can be customized to create emulator $E_{\theta_n}$ specific to UE $n$ taking into account UE $n$'s device hardware configurations and conditions of its environment. Subsequently, this tailored emulator, $E_{\theta_n}$, is distributed to each respective UE. In our work, we extend our proposed \mbox{\texttt{PEAT}} approach to a collaborative, federated model fine-tuning context and establish the Federated Parameter-Efficient Emulator-Assisted Tuning (\mbox{\texttt{FedPEAT}}) framework.

We denote $\mathcal{N}=\{1,2,\ldots,N\}$ and $\mathcal{T}=\{1,2,\ldots,T\}$ as the UE and iteration set for accomplishing the training. At the start of the first iteration, each adapter $A^0_{\theta_n}$ with randomly initialized parameter values is disseminated to UE $n$ where the complete user device model for UE $n$ will be $E^{0}_{\theta_n} \circ A^{0}_{\theta_n}$. And then, the server orchestrator will determine the user selection $\{U_n^t|\forall n\in\mathcal{N}\}$ for participating in the current update, where $U_n^t=0$ signifies non-participation, and $1$ indicates participation. Each selected UE will then carry out model-emulated-assisted fine-tuning with their local dataset $D^t_n$ and update the parameter values of their adapter to produce $A^1_{\theta_n}$ with the assistance of the emulator. Each user device UE $n$ will then upload their adapter parameters to the server for adapter parameter aggregation as follows:
\begin{align}
A^{t+1}_{\phi_g} = \frac{1}{\sum\limits_{n\in\mathcal{N}:U_n^t=1}|D^t_n|}\cdot\sum^{N}_{n=1}(|D^t_n|\cdot A^{t}_{\phi_n}),
\label{eq:R11} 
\end{align}
where $|D^t_n|$ is the size of the data being trained on at UE $n$. The server will then disseminate this global adapter $A^{t+1}_{\phi_g}$. This above-mentioned tuning process proceeds for further iterations until model convergence, or as defined by a specific criterion. This process can be summarized in Algorithm~\ref{alg:FL}.

\section*{\mbox{\texttt{FedPEAT}} Adaptive Control Mechanism}\label{Adaptive}

The \mbox{\texttt{FedPEAT}} framework facilitates the collaborative, federated fine-tuning of models for downstream tasks. However, successful adoption of the framework for downstream task fine-tuning has to be achieved through adaptive control on the hyper-parameters related to the \mbox{\texttt{FedPEAT}} framework, the PEFTs and the FL process. As there are potentially many variables-of-concerns, and diverse scenarios, we designed an adaptive control system which is able to handle the control of multiple variables. To illustrate the \mbox{\texttt{FedPEAT}} with Adaptive Control mechanism, we consider a situation where UEs are moving within a fixed geographic space where the channel gain between the user device and the server changes.

\begin{figure}[!t] 
        \renewcommand{\algorithmicrequire}{\textbf{Initiate:}}
        \renewcommand{\algorithmicensure}{\textbf{Output:}}
        \begin{algorithm}[H]
            \caption{\label{alg:FL} \mbox{\texttt{FedPEAT}} with Adaptive Control}
            \begin{algorithmic}[1]
                \REQUIRE device set $\mathcal{N}=\{1, 2, \ldots, N\}$, initial global adapter $A_{\phi_g}^0$, foundation model $M_{\theta_g}$
                \FOR{iteration $t\in\{1,2...,T\}$}
                    \STATE Adaptive control to decide user selection, emulator sizes, downlink bandwidth, and transmission power resources for every user: $\{(U_n^t, E_{\theta_n}^t, B_n^t, P_n^t)|\forall n\in\mathcal{N}\}$, based on Section~\ref{Adaptive}
                    \STATE Transmit global adapter $A_{\phi_g}^t$, and if the current emulator on user devices needs to be changed, also transmit changed emulators $\{E_{\theta_n}^t|\forall n\in\hspace{-2pt}\mathcal{N}\hspace{-2pt}:{E_{\theta_n}^t\neq E_{\theta_n}^{t-1}}\}$ to devices
                    \FOR{device $n \in \mathcal{N}$ in parallel}
                        \FOR{epoch $\nu=1$ to $V$}
                        \STATE 
                        $A^{t}_{\phi_{n}}[\nu+1] = \text{ModelTuning(}D^t_n, E^t_{\theta_{n}}, A^{t}_{\phi_{n}}[\nu]$).
                        \ENDFOR
                        \STATE $A^{t}_{\phi_{n}}\leftarrow A^{t}_{\phi_{n}}[V]$,
                        \STATE Transmit local $A^{t}_{\phi_{n}}$ to server.
                    \ENDFOR
                \STATE Perform adapter parameter aggregation with equation (\ref{eq:R11}) to obtain $A^{t+1}_{\phi_{g}}$
                \ENDFOR
                    
            \end{algorithmic}
        \end{algorithm}\vspace{-0.8cm}
\end{figure}

\subsection*{Adaptive Control Scenario}
In each iteration, the server orchestration performs key tasks. It begins by selecting users (${U_n^t | n \in \mathcal{N}}$) for participation in the FL process. Next, it determines emulators for each user (${E_{\theta_n}^t | n \in \mathcal{N}}$), arranges downlink bandwidth resources (${B_n^t | n \in \mathcal{N}}$), and allocates downlink transmission power (${P_n^t | n \in \mathcal{N}}$) for effective UE engagement. \textit{Frequency Division Multiple Access} (FDMA) communication technique is adopted to mitigate the interference between UEs associated with different edge servers. 
Similar to the setting in~\cite{lim2021dynamic}, we assume the central server allocates its dedicated bandwidth to the UEs it is associated with.
According to Shannon's formula, the achievable transmission rate of UE $n$ and the central server can be formulated as
\begin{align}
    r_n^t(B_n^t, P_n^t) = B_n^t \log_2 (1+\frac{g_n^t P_n^t}{B_n^t \sigma_0^2}), \label{r_n,m}
\end{align}
where $r_n^t(B_n^t, P_n^t)$ means transmission rate $r_n^t$ is a function of $B_n^t, P_n^t$. $g_n$ is the channel gain between UE $n$ and the central server, with Rician fading being the small-scale fading~\cite{rician}, and $\sigma_0$ is the power spectral density of additive white Gaussian noise.
% In this paper, we assume the bandwidth is equally allocated to all the UEs associated with the edge server.
Note that the total bandwidth the central server can allocate is $B_{\max}$, so we have $\sum_{n\in\mathcal{N}:U_n^t=1} B_n^t \leq B_{\max}$. We also optimize the power allocated by the central server for the downlink transmission of emulator and adapters. Note that the total power the central server can allocate is $P_{\max}$, so we have $\sum_{n\in\mathcal{N}:U_n^t=1} P_n\leq P_{\max}$. As the size of the adapter is small and negligible in the context of emulator assisted-tuning, we assume the adapter is transmitted via a dedicated channel, and ignore the uplink energy and time overhead for adapters. Only if the current emulator is designated for modification, do we proceed to transmit the updated emulator. We introduce an indicator function $\chi[x]$ that equals 1 when event $x$ occurs and 0 otherwise. Then the transmission delay from server to UE $n$ within one iteration can be given as: 
\begin{align}
   d_{n,trans}^t(U_n^t, E_{\theta_n}^t, B_n^t, P_n^t) = U_n^t \times \chi[E_{\theta_n}^t \neq E_{\theta_n}^{t-1}] \times \frac{D(E_{\theta_n}^t)}{r_{n}^t}, \label{t_n_to_m}
\end{align}
where $D(E_{\theta_n}^t$ is the allocated emulator size. Then, the time for one round of local training and model transmission for UE $n$ is $Q_{n}^t=d_{n,comp}^t + d_{n,trans}^t$, where $d_{n,comp}^t$ is the model fine-tuning time taken for iteration $t$ of local training at UE $n$, computed empirically.

Therefore, we formulate the problem as follows:
\begin{subequations}
\label{p1}
    \begin{align}
    \min_{\{(U_n^t, E_{\theta_n}^t, B_n^t, P_n^t)|\forall n\in\mathcal{N}, \forall t\in \mathcal{T}\}}& \Bigg\{\xi_p \cdot \frac{1}{N}\sum_{t=1}^{T}\sum_{n=1}^{N}p^t_n + \xi_f\cdot \sum_{t=1}^{T}\max_{n\in\mathcal{N}}Q^{t}_{n} + \xi_s \cdot \frac{1}{N}\cdot \sum_{t=1}^{T}\sum_{n=1}^{N}\chi[E_{\theta_n}^t \neq E_{\theta_n}^{t-1}] \Bigg\},  \tag{\ref{p1}}\\
    \text{Subject to:  }
    &m^t_n \leq \frac{1}{q}\cdot m^t_{\max,n},~\forall n \in \mathcal{N},\label{p1_2}\\
    &\sum_{t=2}^{T}\chi[E_{\theta_n}^t \neq E_{\theta_n}^{t-1}] \leq T\cdot\frac{1}{c},~\forall n \in \mathcal{N},~\forall t \in \mathcal{T},\label{p1_3}\\
    &\sum_{n\in\mathcal{N}}B_n^t \leq B_{\max},~~ \forall t\in\mathcal{T}, \label{p1_4}\\
    &\sum_{n\in\mathcal{N}}P_n^t \leq P_{\max},~~ \forall t\in\mathcal{T}. \label{p1_5}
    \end{align}
\end{subequations}
\iffalse
\begin{subequations}
\label{p1}
    \begin{align}
    \min_{\{(U_n^t, E_{\theta_n}^t, B_n^t, P_n^t)|\forall n\in\mathcal{N}, \forall t\in \mathcal{T}\}}& \Bigg\{\xi_p \cdot \frac{1}{N}\sum_{t=1}^{T}\sum_{n=1}^{N}p^t_n + \xi_f\cdot \sum_{t=1}^{T}\max_{n\in\mathcal{N}}Q^{t}_{n} + \xi_b \cdot \frac{1}{N}\cdot \sum_{t=1}^{T}\sum_{n=1}^{N}m^t_n + \xi_s \cdot \frac{1}{N}\cdot \sum_{t=1}^{T}\sum_{n=1}^{N}\chi[E_{\theta_n}^t \neq E_{\theta_n}^{t-1}] \Bigg\},  \tag{\ref{p1}}\\
    \text{Subject to:  }
    &m^t_n \leq \frac{1}{d}\cdot M,~\forall n \in \mathcal{N},\label{p1_2}\\
    &\sum_{t=2}^{T}\chi[E_{\theta_n}^t \neq E_{\theta_n}^{t-1}] \leq T\cdot\frac{1}{c},~\forall n \in \mathcal{N},~\forall t \in \mathcal{T},\label{p1_3}\\
    &\sum_{n\in\mathcal{N}}B_n^t \leq B_{\max},~~ \forall t\in\mathcal{T}, \label{p1_4}\\
    &\sum_{n\in\mathcal{N}}P_n^t \leq P_{\max},~~ \forall t\in\mathcal{T}. \label{p1_5}
    \end{align}
\end{subequations}
\fi

% \textcolor{red}{need to add another variable for representing the emulator switch, something like $\chi[E_{\theta_n}^t\neq E_{\theta_n}^{t-1}]$, where $\chi[A]$ equals 1 if A happens, 0 otherwise. And we multiply $\chi[]$ to the wireless communication cost, making the problem formulation more clear.}

In the objective function~(\ref{p1}) above, $p^t_n$ stands for the perplexity score achieved by UE $n$ at iteration $t$, where it is a performance measure for how well a language model predicts a set of data, $Q^t_n$ stands for the log of total time taken for a single round of adapter and emulator transmission. $\chi[E_{\theta_n}^t \neq E_{\theta_n}^{t-1}]$ represents the emulator exchange count. $\xi_p$, $\xi_f$, and $\xi_s$ stand for the weight balancing parameters for these objectives. $m^t_n$ represents the memory space taken for the model assigned to UE $n$, while $m^t_{\max,n}$ represents the available memory capacity of device $n$ at iteration $t$. $c$ and $q$ are numerical constants. Constraint (\ref{p1_2}) ensures that the total memory consumed for any device in each round falls well below a predefined fraction of its memory capacity. Constraint (\ref{p1_3}) prevents excessive emulator switch counts to reduce transmission costs. Constraint (\ref{p1_4}), (\ref{p1_5}) are the limits of total bandwidth and power resources from the server. Essentially, the objective function~(\ref{p1}) aims to minimize the sum of perplexity scores across $T$ iterations which is synonymous with achieving a quicker rate of model tuning convergence, minimizing the maximum training time amongst all devices for the federated fine-tuning process, and emulator exchange count, via optimizing the emulator compression parameter, device selection vector, bandwidth selection vector, and downlink power selection vector. For the sake of simplicity in our demonstration,  we use $\frac{1}{N}\sum_{n=1}^{N}p^t_n$ as an estimate of global $p^t$. The rationale behind such a formulation is to expedite model convergence, all while maintaining the constraint of the maximum total transmission and computation delay among UEs. Additionally, this approach ensures that the sizes of both the emulator and adapters remain within a practical fraction of the local devices' memory capacity.

\subsection*{Deep reinforcement learning approach}\label{RLapproach}
We have devised a deep reinforcement learning approach as our driver behind our adaptive control mechanism to tackle our proposed problem as the problem is highly sequential and is a mixed-integer non-linear programming problem.
\iffalse
The ingenious configuration of the state, action spaces, and reward function plays a crucial role in the effective integration of reinforcement learning techniques. In the upcoming sections, we will delve into a detailed examination of these three components in our approach. We choose a multi-output single-agent deep reinforcement learning approach over a multi-agent approach for simplicity of demonstration. The choice of deep reinforcement learning agent type and structure is arbitrary.
\fi

\subsubsection*{State}
To effectively execute the \mbox{\texttt{FedPEAT}} approach,  we included the following variables within the state: (1) user device-server channel gain $g_n^t$ which is required for the computation of $r^t_{n}$, (2) user device available memory capacity $m^t_n$, (3) \mbox{\texttt{FedPEAT}} UE $n$ emulator exchange count $\chi[E_{\theta_n}^t \neq E_{\theta_n}^{t-1}]$ which keeps track of the number of times UE $n$ has undergone emulator exchange. Output information from each successive actor branch is appended to the state to be fed into the next actor branch as shown in Figure~\ref{fig:architecture}. These additional information include user-selection $U^t_n$, bandwidth selection $B^t_n$ and power selection $P^t_n$ at the current time step.

\subsubsection*{Action}
In this study, we have 4 actions to include in the agent action space: (1) UE selection vector $\{U_n^t|\forall n\in\mathcal{N}\}$, (2) downlink bandwidth selection vector $\{B_n^t|\forall n\in\mathcal{N}\}$, (3) downlink power selection vector $\{P_n^t|\forall n\in\mathcal{N}\}$ (4) choice of emulator compression parameter $\{E_{\theta_n}^t|\forall n\in\mathcal{N}\}$ for each device, stored in a vector.

\subsubsection*{Reward}
We formulate our reward function as per our objective function, where we assign our reinforcement learning agent the reward as follows in each iteration:
\iffalse
\begin{align}
    R = &-\omega_p\frac{1}{TN}\sum_{t=1}^{T}\sum_{n=1}^{N}p^t_n - \omega_f\cdot\frac{1}{T} \sum_{t=1}^{T}\max_{n\in\mathcal{N}}Q^{t}_{n} - \omega_b \cdot \frac{1}{T N}\cdot \sum_{t=1}^{T}\sum_{n=1}^{N}m^t_n - \omega_s \cdot \frac{1}{TN}\cdot \sum_{t=1}^{T}\sum_{n=1}^{N}\chi[E_{\theta_n}^t \neq E_{\theta_n}^{t-1}]\label{rewardfunction}
\end{align}
\fi
\begin{align}
    R^t_d = - \xi_f\cdot\frac{1}{T} \sum_{t=1}^{T}\max_{n\in\mathcal{N}}Q^{t}_{n},~~~
    R^t_p = -\xi_p\frac{1}{TN}\sum_{t=1}^{T}\sum_{n=1}^{N}p^t_n,~~~
    R^t_s = -\xi_s \cdot \frac{1}{TN}\cdot \sum_{t=1}^{T}\sum_{n=1}^{N}\chi[E_{\theta_n}^t \neq E_{\theta_n}^{t-1}].\label{rewardset}
\end{align}
In addition, we assign the agent very large penalties $\varkappa$ when (1) the memory size of emulator $E_{\theta_n}^t$ and adapter $A_{\phi_n}^t$ exceeds an allowable fraction of the local device $n$'s memory capacity, in accordance to constraint~(\ref{p1_2}), and (2) the emulator exchange count $\chi[E_{\theta_n}^t \neq E_{\theta_n}^{t-1}]$ exceeds a given fraction of the total iteration, in accordance to constraint~(\ref{p1_3}).

\subsubsection*{Reinforcement Learning Algorithm}
We adopted the Proximal Policy Optimization (PPO) algorithm, developed by OpenAI~\cite{PPO}, which stands as an advancement over traditional policy gradient algorithms. In the domain of sequential problems such as reinforcement learning, even minor adjustments to parameters can have a profound impact on performance, making parameter fine-tuning a challenging endeavor. PPO tackles the issue of delicate and noisy advantage estimates by implementing a cautious approach. It incorporates a Kullback–Leibler (KL) divergence penalty to regulate policy adjustments. Furthermore, PPO makes use of an importance sampling technique~\cite{kahn1951estimation} by employing asynchronous policies for training and data collection, enhancing overall efficiency. The loss function for the Actor is formally defined as follows~\cite{PPO}:
\begin{align}
    L^{CLIP}(\varphi) = \mathbb{E}^{t}\left[\min\left(\mathfrak{r}^{t}({\varphi})\varpi^{t},\text{clip}\left(\mathfrak{r}^{t}(\varphi),1-\epsilon,1+\epsilon\right)\varpi^{t}\right)\right].\nonumber
\end{align}
In this context, $\varphi$ represents the policy. $\mathbb{E}^{t}$ signifies empirical expectations over the trajectory.
$\mathfrak{r}^{t}$ represents the ratio of the current policy to the old policy.
$\varpi^{t}$ denotes the estimated advantage at time $t$ and $\epsilon$ denotes the clip value. This clipping mechanism acts as a safeguard, preventing significant bias and ensuring that the policy remains within a trusted range.

\begin{figure*}[htbp] 
\centering
\setlength{\abovecaptionskip}{-0.1cm}
\includegraphics[width=0.8\linewidth]{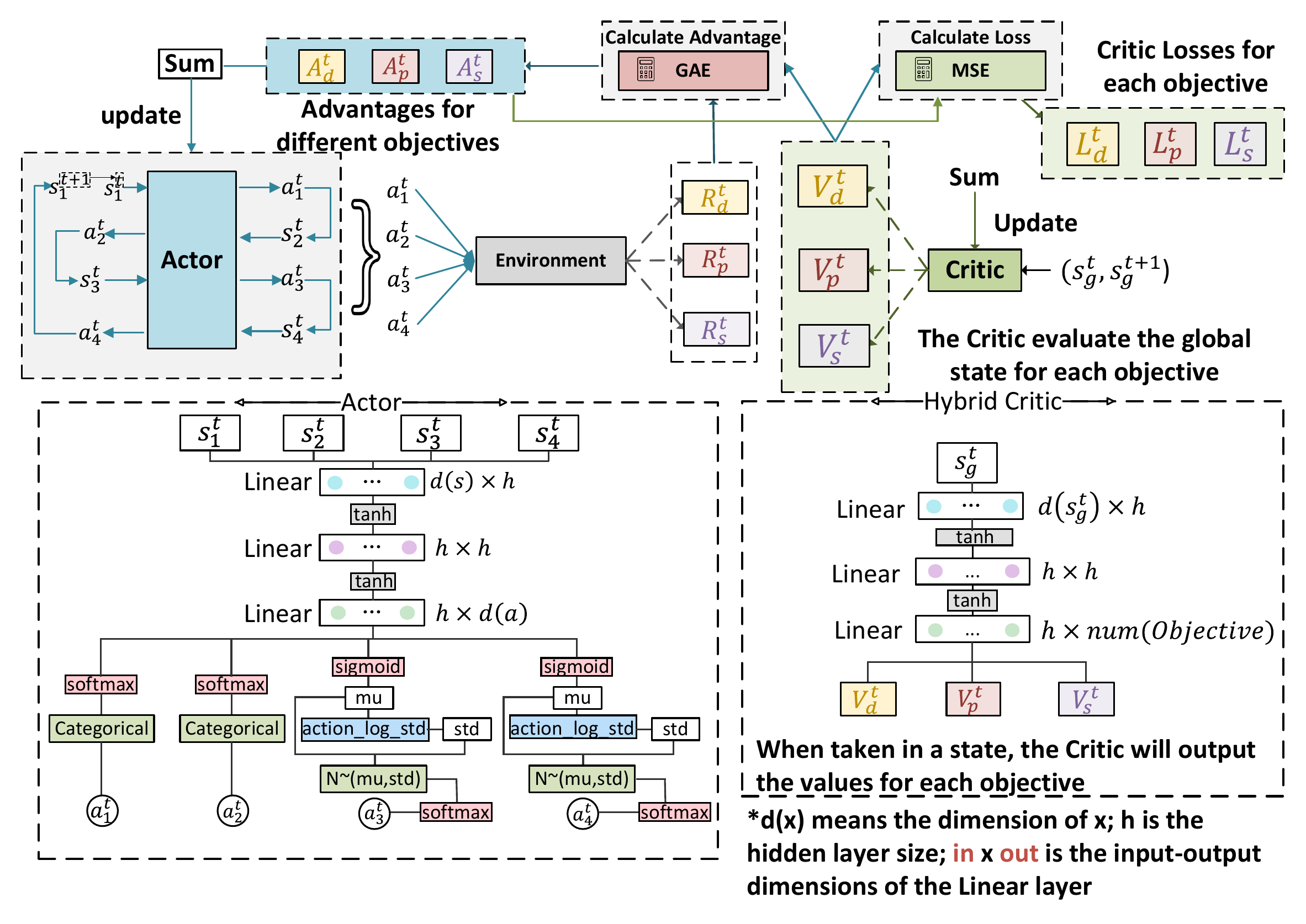}
\caption{Our proposed SABPPO algorithm and architecture. Figure illustrates the underlying actor and critic architecture, their interaction with the environment and model update process.}\vspace{-10pt}
\label{fig:architecture}
\end{figure*}

\begin{figure}[!t] 
        \vspace{-0cm}
        \renewcommand{\algorithmicrequire}{\textbf{Initiate:}}
        \renewcommand{\algorithmicensure}{\textbf{Output:}}
        \begin{algorithm}[H]
            \caption{\label{alg:PPO}SABPPO adaptive control algorithm}
            \begin{algorithmic}[1]
                \REQUIRE critic parameter $\phi$, critic target parameter $\phi'$, $\mathcal{A}$ actor parameter $\varphi$ and data-sampling parameter $\varphi^{'}$, initialize state $s^t_g=s^1_g$;
                \FOR{iteration = $1,2,...$}
                    \FOR{action = $1,2,...$}
                        \STATE $\mathcal{A_{\text{action}}}$ choose action vector $a^t_{\text{action}}$ according to $\pi_{\varphi'}(a^t_\text{action}|s^{t}_\text{action})$, based on SABPPO section.
                        \STATE $s^{t}_{\text{action}+1}$ = concatenate$\{s^{t}_\text{action}, a^t_\text{action}\}$.
                    \ENDFOR
                    \STATE Get $R^{{t}}_d$, $R^{{t}}_p$, $R^{{t}}_s$ based on equation (\ref{rewardset}) and next state $s^{t+1}_1$ from the environment.
                    \STATE Collect trajectories: \\$\tau$=$\{s^t_g,\mathfrak{a}^{{t}}, s^{t+1}_g,R^{{t}}_d, R^{{t}}_p, R^{{t}}_s$\} iteratively till end of episode.
                    \STATE $s^t_g  \leftarrow s^{t+1}_g$;
                    \STATE Compute advantage $\varpi^{{\textcolor{blue}{t}}}$ based on equation (\ref{advantage})
                    % \STATE Compute target values \{$V^{\textcolor{blue}{R},t}_{\textcolor{blue}{\varsigma}},V^{\textcolor{blue}{\vartheta},t}_{\textcolor{blue}{\varsigma}},V^{\textcolor{blue}{\rho},t}_{\textcolor{blue}{\varsigma}}$\}
                    \FOR{$o$ = $1,2,...,O$}
                        \STATE Group trajectories into batches
                        \FOR{each batch}
                            \STATE Compute gradient for actor:
                            $\triangledown \varphi$ based on equation (\ref{eq:actorloss})
                            \STATE Apply gradient ascent on $\varphi$ using $\triangledown \varphi$
                            \STATE Update critic model through back-propagation of loss using equation~based on (\ref{eq:criticloss})
                        \ENDFOR
                        \STATE Update parameters of critic target network $\phi'$ with parameters of critic network $\phi$, every $C$ number of iterations, where $C$ denotes the interval for critic parameter update;
                    \ENDFOR
                    
                \ENDFOR
            \end{algorithmic}
        \end{algorithm}
\vspace{0.0cm}
\end{figure}

\subsection*{Single-Agent Action Branching Proximal Policy Optimization (SABPPO)\label{SAB-PPO}}
\iffalse
\textcolor{blue}{(1) Discuss the need to handle multiple action/high-dimension action when we are scaling the adaptive control\\
(2) Propose a new Action Branching architecture Based on previous work, and, why is this approach good?\\
(3) what is our improvement and why we do this?\\
(4) go into detail of the algorithm/architecture\\
(5) include figure for SABPPO}\\
(6) include an algorithm for the DRL training\\
\fi

To ensure scalable federated fine-tuning of foundation models system, multiple variables require optimization. However, as the number of optimization variables increase, the number of actions that need to be explicitly represented grows exponentially with increasing action dimensionality, where the total number of actions equates to $\prod^{\mathfrak{D}}_{\mathfrak{d=1}}|\mathfrak{a}_\mathfrak{d}|$, where $\mathfrak{D}$ is the total number of action dimensions, $\mathfrak{a}_\mathfrak{d}$ is the action space of action $\mathfrak{d}$. However, traditional deep reinforcement learning architectures do not handle the exponentially growing action dimension well. We propose a novel \textbf{Single-Agent Action Branching Proximal Policy Optimization (SABPPO)} algorithm which is inspired by the action branching approaches proposed by~\cite{tavakoli2018action}. The SABPPO architecture builds on state-of-the-art Proximal Policy Optimization algorithm~\cite{PPO} and distributes the representation of the action controllers across individual network branches, meanwhile, maintaining a shared decision module among them to encode a latent representation of the input and help with the coordination of the branches. This proposed approach enables the linear growth of the total number of network outputs with increasing action dimensionality as opposed to the combinatorial growth in current discrete-action algorithms. SABPPO extends the PPO architecture with a single critic and actor. In each FL iteration, the actor's user-selection branch takes the state $s^t_1$ as input, producing user-selection information for concatenation with the state to form $s^t_2$. This concatenated state is then input to the actor's bandwidth-selection branch, generating bandwidth selection information for concatenation with the state to form $s^t_3$. The same process occurs with the power-selection branch, producing power selection information for concatenation with the state to form $s^t_4$. Lastly, the concatenated state is fed into the actor's emulator-compression selection branch, yielding emulator compression selection information (shown in Figure~\ref{fig:architecture}). The SABPPO actor is updated as follow:
\begin{align}
    &\Delta\varphi \hspace{-2pt}=\hspace{-2pt} \mathbb{E}^t[\nabla_{\varphi} \min\{\mathfrak{r}^t(\varphi)\varpi^t, \text{clip}(\mathfrak{r}^t(\varphi), 1\hspace{-2pt}-\hspace{-2pt}\epsilon, 1\hspace{-2pt}+\hspace{-2pt}\epsilon)\varpi^t\}], \label{eq:actorloss}
\end{align}
while the SABPPO critic is updated as follows:
\begin{align}
    &L^t(\phi) = [V_\phi(s^t_g)-(\varpi^t+V_{\phi'}(s^t_g))]^2. \label{eq:criticloss}
\end{align}
$s^t_g$ is the global state, which is the concatenation of all states, $V$ is the state-value function and $\phi$ and $\phi'$ are the state-value function parameter and target state-value function parameter, respectively. Here, $\mathfrak{r}^t(\theta)$ represents the ratio between the two policies:
$\mathfrak{r}^t(\theta)\hspace{-2pt}=\hspace{-2pt}\frac{\pi_\theta(a^t_1,a^t_2,a^t_3,a^t_4|s^t_g)}{\pi_{{\theta'}}(a^t_1,a^t_2,a^t_3,a^t_4|s^t_g)}$. And the advantage function $\varpi^t$ is calculated via Generalized Advantage Estimation (GAE)~\cite{schulman2015high}:
\begin{align}
    &\varpi^t = \delta^t + (\gamma\lambda)\delta^{t+1}+...+(\gamma\lambda)^{\bar{T}-1}\delta^{t+\bar{T}-1}, ~~~\text{where}~~\delta^t=R^t+\gamma V_{\phi'}(s^{t+1}_g)-V_{\phi'}(s^{t}_g)\label{advantage},
\end{align}
$\bar{T}$ is the trajectory segment, $\lambda$ is the trace decay parameter and $\gamma$ is the discount rate.

\begin{figure*}[htbp]
\centering
\subfigtopskip=2pt
\subfigbottomskip=2pt
\subfigure[FL vs FedPEAT (Delay).]{
\begin{minipage}[t]{0.25\linewidth}
\centering
\includegraphics[width=0.7\linewidth]{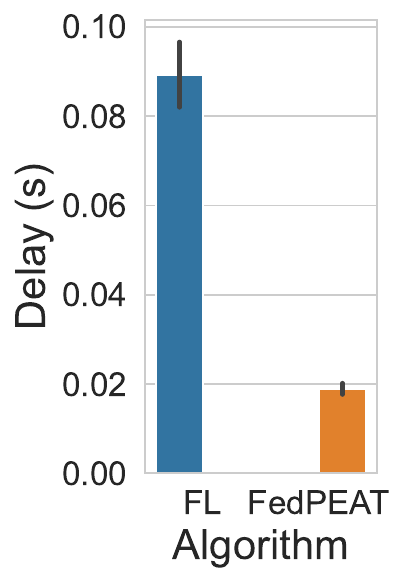}
\label{fig:fl_delay}
\vspace{-10mm}
\end{minipage}
}%
\subfigure[FL vs FedPEAT (Emulator Switch)]{
\begin{minipage}[t]{0.25\linewidth}
\centering
\includegraphics[width=0.67\linewidth]{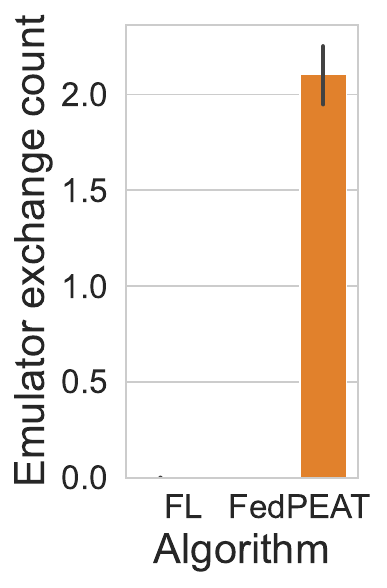}
\label{fig:fl_emuxchange}
\vspace{-10mm}
\end{minipage}%
}%
\subfigure[FL vs FedPEAT (Perplexity).]{
\begin{minipage}[t]{0.25\linewidth}
\centering
\includegraphics[width=0.65\linewidth]{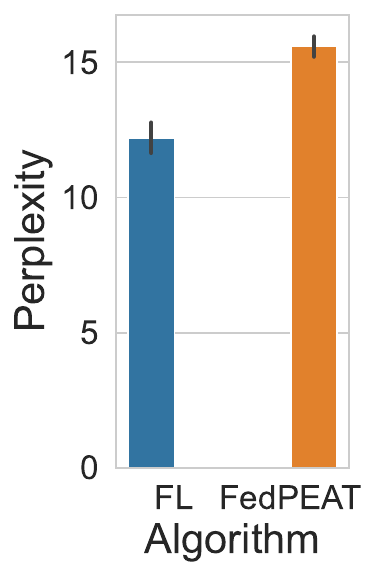}
\label{fig:fl_perplexity}
\vspace{-10mm}
\end{minipage}%
}%
\subfigure[Algorithm training time.]{
\begin{minipage}[t]{0.25\linewidth}
\centering
\includegraphics[width=0.78\linewidth]{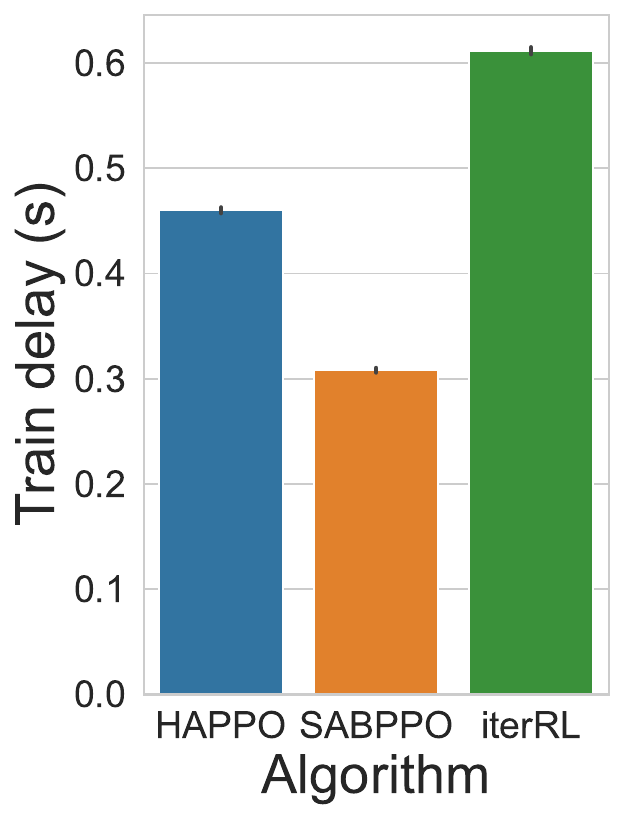}
\label{fig:traintime}
\vspace{-10mm}
\end{minipage}
}%

\subfigure[log(Delay)]{
\begin{minipage}[t]{0.25\linewidth}
\centering
\includegraphics[width=1\linewidth]{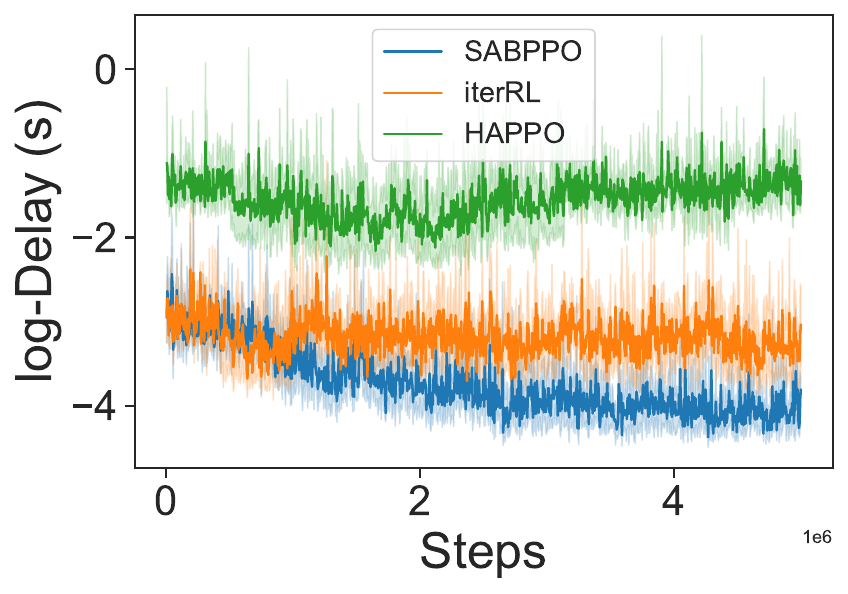}
\label{fig:fed_delay}
\vspace{-10mm}
\end{minipage}%
}%
\subfigure[Emulator exchange count]{
\begin{minipage}[t]{0.25\linewidth}
\centering
\includegraphics[width=1\linewidth]{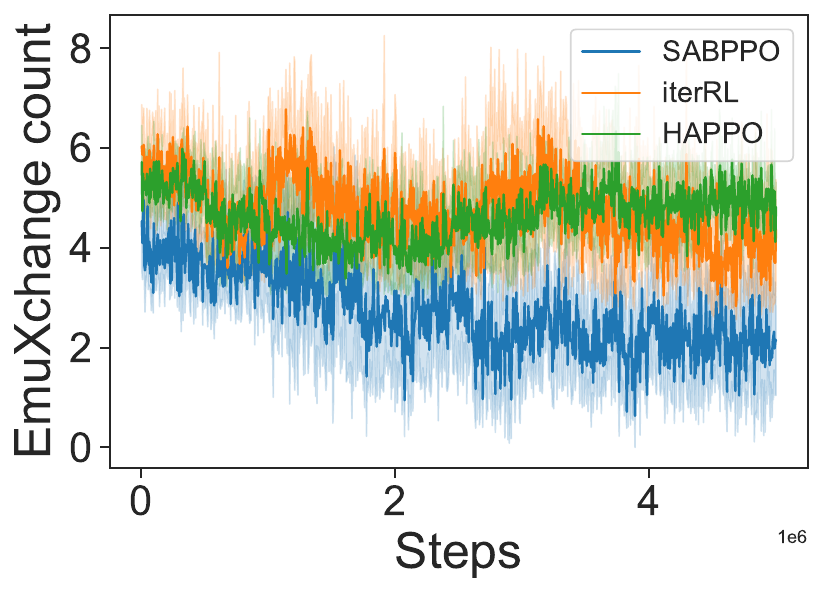}
\label{fig:fed_emuxchange}
\vspace{-10mm}
\end{minipage}%
}%
\subfigure[Perplexity]{
\begin{minipage}[t]{0.25\linewidth}
\centering
\includegraphics[width=1\linewidth]{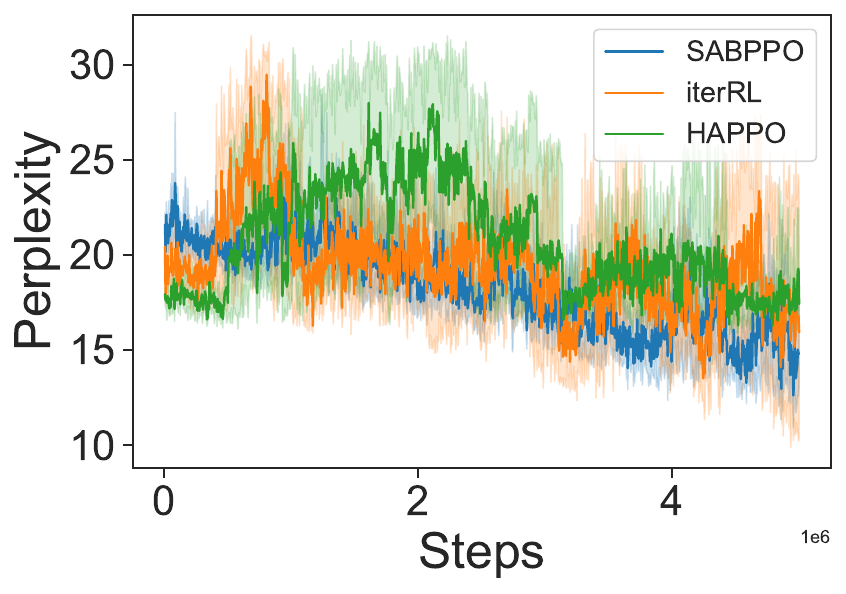}
\label{fig:fed_perplexity}
\vspace{-10mm}
\end{minipage}%
}%
\subfigure[Reward]{
\begin{minipage}[t]{0.25\linewidth}
\centering
\includegraphics[width=1\linewidth]{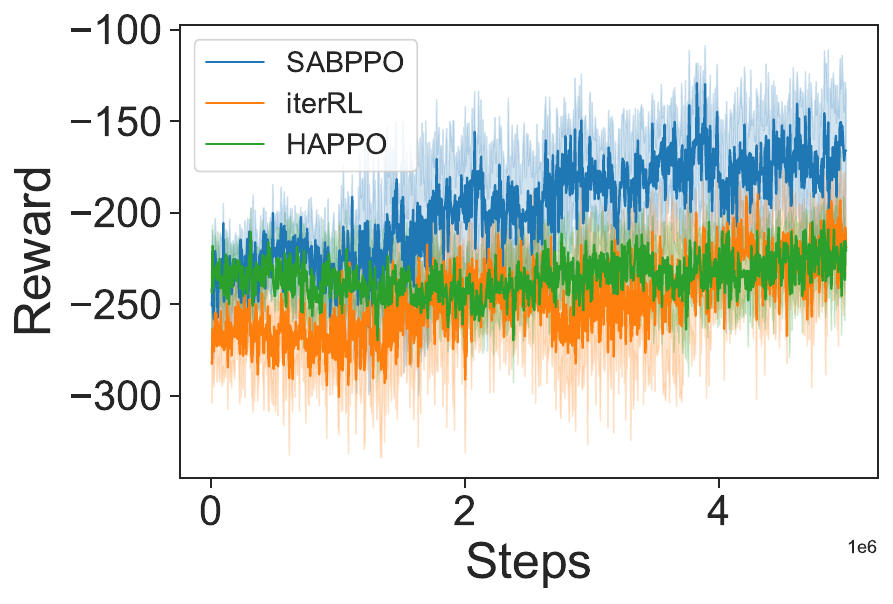}
\label{fig:fed_reward}
\vspace{-10mm}
\end{minipage}%
}%

\caption{Comparison between FL and \mbox{\texttt{FedPEAT}}, and Comparison between Adaptive control algorithms. Figure \ref{fig:fl_delay}, \ref{fig:fl_emuxchange}, \ref{fig:fl_perplexity} illustrates the performance difference between FL and \mbox{\texttt{FedPEAT}} with regards to delay, emulator exchange count, and perplexity, respectively. Figure~\ref{fig:traintime} illustrates the time taken for model training for each adaptive control algorithm. Figure \ref{fig:fed_delay}, \ref{fig:fed_emuxchange}, \ref{fig:fed_perplexity}, \ref{fig:fed_reward} illustrate the performance of each adaptive control algorithm in across the training process, in terms of log(delay), emulator exchange count, perplexity, and reward.}
\label{fig_metric}
\vspace{-0.5cm}
\end{figure*}

\section*{Numerical Experiments}
\iffalse
In this section, we will delve into our experimental configurations and offer a comprehensive analysis of the obtained results.
\fi

% The configuration settings are designed for simplicity of demonstration and are arbitrary. We consider the scenario where the server has data and partakes in UE-server collaborative federated learning.

\subsection*{Experiment configuration}
We substantiate our study with several experiments by showing that the \mbox{\texttt{FedPEAT}} framework with adaptive control works, and compare our proposed framework against Federated full model fine-tuning (Fed-FT). To simplify our workflow and for the ease of demonstration, we utilized numerical solutions from the works by~\cite{xiao2023offsite} to facilitate our experiment. We utilized the \textit{OPT-1.3B}~\cite{radford2019language} large language model as the foundation model, which has 1208 million parameters and is of approximately 2.63 gigabytes (GB) in storage memory. We utilized the layer-drop approach~\cite{zhang2020accelerating} for the emulator compression. We adopted the perplexity-layer drop retention numerical solution from the works by~\cite{xiao2023offsite} and established the function to be approximated by $P = 25.2\varrho^2 - 43.1\varrho + 31.9$ for $0 < \varrho \leq 1$, with an $R^2$ score of $0.97$, where $\varrho$ stands for the layer drop retention ratio and varies from 0 to 1. We also adopted perplexity improvements of using LoRA  from the works by~\cite{xiao2023offsite}, and establish the perplexity improvement upon application of LoRA to be $-0.78$. We designed the trainable layers of the adapter to be 2 layers at both top and bottom layers of the neural network. We assume model storage memory usage to follow a linear relationship with the number of parameters of the model. We set $T$ which is the number of federated fine-tuning rounds in an episode to be 100. We consider a scenario with 1 main server and 10 user devices. In each round of federated fine-tuning, our adaptive control orchestrator selects 5 devices for fine-tuning. To facilitate the collaboratively training scenario, the server holds $30\%$ of the total data to be trained. We set our large penalty $\varkappa$ to be -50.

As we consider the communications to be over 6G networks, we assign the bandwidth $B$ to be selected from a range between $7$ and $20$ Ghz and noise $\sigma^2$ to be $-174$ dBm. We initialize and constrain main server power output to $(0.0,15.0)$ Watt. User-device channel gain are calculated based on path-loss~\cite{erceg1999empirically} and user distance from the server. $\xi_p$, $\xi_f$, and $\xi_s$ are set to 5, -10, and 25, respectively, and these numbers are empirically derived with the aim of balancing the variables in the objective function. We adopt the ADAM optimizer~\cite{kingma2014adam} for the algorithms implemented in our study. The models are trained for 5,000,000 steps and evaluated at every 5000 steps.

\subsection*{Experiment results}

Our findings reveal that \mbox{\texttt{FedPEAT}} with adaptive control outperforms Fed-FT significantly in terms of both communication and computation delays. Specifically, in a single round of computation and communication, Fed-FT demonstrates a delay 4.60$\times$ longer than that of \mbox{\texttt{FedPEAT}} with adaptive control, as illustrated in Figure~\ref{fig:fl_delay}. This notable improvement by \mbox{\texttt{FedPEAT}} with adaptive control is achieved despite the need for an emulator exchange, where emulators are exchanged 2.10 $\times$ on average in each 10 iterations (Figure~\ref{fig:fl_emuxchange}), a process not required in Fed-FT. The ability of \mbox{\texttt{FedPEAT}} with adaptive control to mitigate delays underscores its effectiveness in enhancing the efficiency of FL systems, even when accounting for additional emulator exchanges. However, it is essential to note that \mbox{\texttt{FedPEAT}} with Adaptive control exhibits a perplexity score 3.49 points higher than that of Fed-FL.

Subsequently, we extend our analysis to compare our proposed SABPPO adaptive control algorithm with baseline algorithms, namely iterative Reinforcement Learning Agents (iterRL) and Heterogeneous Action Proximal Policy Optimization (HAPPO). IterRL employs independent actors and critics, while HAPPO, based on Centralized Training and Decentralized Execution (CTDE)~\cite{lowe2017multi}, features three separate actors sharing a single critic model. Our experimental results, as depicted in Figure~\ref{fig:fed_reward}, showcase the superiority of the SABPPO algorithm in terms of model convergence and reward. Specifically, SABPPO achieves a reward of -168, outperforming HAPPO and iterRL, which attain -225 and -214, respectively, after 5,000,000 training steps. This superior performance is corroborated by lower log(delay) values (Figure~\ref{fig:fed_delay}) of -4.02 for SABPPO compared to -1.33 and -3.27 for HAPPO and iterRL, respectively. Additionally, SABPPO exhibits fewer emulator exchanges (2.10$\times$) compared to HAPPO (4.91$\times$) and iterRL (3.95$\times$) (Figure~\ref{fig:fed_emuxchange}), as well as lower perplexity scores (15.03 compared to 17.86 and 16.60 for HAPPO and iterRL, respectively), as seen in Figure~\ref{fig:fed_perplexity}).

Furthermore, the SABPPO framework demonstrates a significantly shorter delay in model training (0.308 seconds) compared to HAPPO (0.534 seconds) and iterRL (0.549 seconds), as highlighted in our results. These findings collectively underscore the efficiency gains achieved by the SABPPO algorithm across various performance metrics when compared to baseline algorithms in federated learning scenarios.

\section*{Discussion and Conclusion}
In summary, the deployment and refinement of large foundation models present multifaceted challenges, necessitating solutions that address collaborative training, model ownership, and computational constraints to fully unlock their potential. In response to these challenges, we extend the offsite tuning paradigm to introduce Emulator-Assisted Tuning (\mbox{\texttt{EAT}}) and integrate it with Parameter-Efficient Fine-Tuning (PEFT), resulting in the creation of Parameter-Efficient Emulator-Assisted Tuning (\mbox{\texttt{PEAT}}). This novel approach is further extended to the FL domain, resulting in Federated Parameter-Efficient Emulator-Assisted Tuning (\mbox{\texttt{FedPEAT}}).

Our proposed \mbox{\texttt{FedPEAT}} framework, featuring adaptive control and a unique fusion of adapters and emulators, represents a pioneering avenue for advancing model privacy and optimizing memory-efficient downstream federated fine-tuning. Adapters, endowed with trainable neural network parameters, tailor models for specific tasks, while emulators offer compressed, fixed-parameter representations. This innovative approach not only addresses concerns regarding model privacy by eliminating the need to transmit complete models to edge devices but also substantially enhances memory and computational efficiency. Its adaptability to diverse neural network architectures is complemented by an adaptive control mechanism, employing deep reinforcement learning to optimize critical hyper-parameters, thus ensuring efficient resource orchestration.

In the broader context, our \mbox{\texttt{FedPEAT}} framework, empowered by the SABPPO-Adaptive control optimizer, facilitates Federated fine-tuning by leveraging Parameter-Efficient Fine-Tuning (PEFT) and Emulator-Assisted Tuning (EAT) methodologies. This framework upholds user data privacy through Federated Learning and protects model owner intellectual property (IP) through \mbox{\texttt{EAT}}. Furthermore, our experimental results demonstrate that \mbox{\texttt{FedPEAT}} with Adaptive control significantly outperforms traditional Federated Learning (Fed-FT) in terms of communication and computation efficiency as the use of \mbox{\texttt{FedPEAT}} reduces the foundation model memory footprint and number of parameters to tune. This efficiency gain enables the inclusion of low-resource devices in federated fine-tuning of foundation models. While \mbox{\texttt{FedPEAT}} with adaptive control exhibits a slightly higher perplexity score compared to Fed-FT, the marginal discrepancy in performance is overshadowed by the substantial reduction in communication and computation overhead. Notably, the adaptive control mechanism can be fine-tuned to prioritize higher perplexity scores should specific preferences or requirements dictate such adjustments.

Moreover, our investigation reveals that the training time required for the proposed SABPPO adaptive control optimizer is significantly lower than that achieved by the HAPPO and iterRL algorithms. This efficiency gain is attributed to the streamlined training process of a single actor and a single critic in SABPPO, as opposed to the more resource-intensive training requirements of multiple actors and critics in HAPPO and iterRL. This reduction in the number of neural networks trained concurrently contributes significantly to the observed decrease in training time. In conclusion, our comprehensive framework, \mbox{\texttt{FedPEAT}} with adaptive control, stands as a pioneering solution, providing a nuanced balance between model performance, privacy, and resource efficiency in the complex landscape of federated learning and large model fine-tuning.
\bibliography{sample}

\begin{thebibliography}{10}
\urlstyle{rm}
\expandafter\ifx\csname url\endcsname\relax
  \def\url#1{\texttt{#1}}\fi
\expandafter\ifx\csname urlprefix\endcsname\relax\def\urlprefix{URL }\fi
\expandafter\ifx\csname doiprefix\endcsname\relax\def\doiprefix{DOI: }\fi
\providecommand{\bibinfo}[2]{#2}
\providecommand{\eprint}[2][]{\url{#2}}

\bibitem{brown2020language}
\bibinfo{author}{Brown, T.} \emph{et~al.}
\newblock \bibinfo{journal}{\bibinfo{title}{Language models are few-shot learners}}.
\newblock {\emph{\JournalTitle{Advances in Neural Information Processing Systems}}} \textbf{\bibinfo{volume}{33}}, \bibinfo{pages}{1877--1901} (\bibinfo{year}{2020}).

\bibitem{radford2019language}
\bibinfo{author}{Radford, A.} \emph{et~al.}
\newblock \bibinfo{journal}{\bibinfo{title}{Language models are unsupervised multitask learners}}.
\newblock {\emph{\JournalTitle{OpenAI blog}}} \textbf{\bibinfo{volume}{1}}, \bibinfo{pages}{9} (\bibinfo{year}{2019}).

\bibitem{devlin2018bert}
\bibinfo{author}{Devlin, J.}, \bibinfo{author}{Chang, M.-W.}, \bibinfo{author}{Lee, K.} \& \bibinfo{author}{Toutanova, K.}
\newblock \bibinfo{journal}{\bibinfo{title}{{BERT}: Pre-training of deep bidirectional transformers for language understanding}}.
\newblock {\emph{\JournalTitle{arXiv preprint arXiv:1810.04805}}}  (\bibinfo{year}{2018}).

\bibitem{wei2021finetuned}
\bibinfo{author}{Wei, J.} \emph{et~al.}
\newblock \bibinfo{journal}{\bibinfo{title}{Finetuned language models are zero-shot learners}}.
\newblock {\emph{\JournalTitle{arXiv preprint arXiv:2109.01652}}}  (\bibinfo{year}{2021}).

\bibitem{muennighoff2022crosslingual}
\bibinfo{author}{Muennighoff, N.} \emph{et~al.}
\newblock \bibinfo{journal}{\bibinfo{title}{Crosslingual generalization through multitask finetuning}}.
\newblock {\emph{\JournalTitle{arXiv preprint arXiv:2211.01786}}}  (\bibinfo{year}{2022}).

\bibitem{letaief2019roadmap}
\bibinfo{author}{Letaief, K.~B.}, \bibinfo{author}{Chen, W.}, \bibinfo{author}{Shi, Y.}, \bibinfo{author}{Zhang, J.} \& \bibinfo{author}{Zhang, Y.-J.~A.}
\newblock \bibinfo{journal}{\bibinfo{title}{The roadmap to 6g: Ai empowered wireless networks}}.
\newblock {\emph{\JournalTitle{IEEE communications magazine}}} \textbf{\bibinfo{volume}{57}}, \bibinfo{pages}{84--90} (\bibinfo{year}{2019}).

\bibitem{mcmahan2017communication}
\bibinfo{author}{McMahan, B.}, \bibinfo{author}{Moore, E.}, \bibinfo{author}{Ramage, D.}, \bibinfo{author}{Hampson, S.} \& \bibinfo{author}{y~Arcas, B.~A.}
\newblock \bibinfo{title}{Communication-efficient learning of deep networks from decentralized data}.
\newblock In \emph{\bibinfo{booktitle}{Artificial Intelligence and Statistics}}, \bibinfo{pages}{1273--1282} (\bibinfo{organization}{PMLR}, \bibinfo{year}{2017}).

\bibitem{konevcny2016federated}
\bibinfo{author}{Kone{\v{c}}n{\`y}, J.} \emph{et~al.}
\newblock \bibinfo{journal}{\bibinfo{title}{Federated learning: Strategies for improving communication efficiency}}.
\newblock {\emph{\JournalTitle{arXiv preprint arXiv:1610.05492}}}  (\bibinfo{year}{2016}).

\bibitem{bonawitz2019towards}
\bibinfo{author}{Bonawitz, K.} \emph{et~al.}
\newblock \bibinfo{journal}{\bibinfo{title}{Towards federated learning at scale: System design}}.
\newblock {\emph{\JournalTitle{Proceedings of Machine Learning and Systems}}} \textbf{\bibinfo{volume}{1}}, \bibinfo{pages}{374--388} (\bibinfo{year}{2019}).

\bibitem{rebuffi2017learning}
\bibinfo{author}{Rebuffi, S.-A.}, \bibinfo{author}{Bilen, H.} \& \bibinfo{author}{Vedaldi, A.}
\newblock \bibinfo{journal}{\bibinfo{title}{Learning multiple visual domains with residual adapters}}.
\newblock {\emph{\JournalTitle{Advances in Neural Information Processing Systems}}} \textbf{\bibinfo{volume}{30}} (\bibinfo{year}{2017}).

\bibitem{he2021towards}
\bibinfo{author}{He, J.}, \bibinfo{author}{Zhou, C.}, \bibinfo{author}{Ma, X.}, \bibinfo{author}{Berg-Kirkpatrick, T.} \& \bibinfo{author}{Neubig, G.}
\newblock \bibinfo{journal}{\bibinfo{title}{Towards a unified view of parameter-efficient transfer learning}}.
\newblock {\emph{\JournalTitle{arXiv preprint arXiv:2110.04366}}}  (\bibinfo{year}{2021}).

\bibitem{houlsby2019parameter}
\bibinfo{author}{Houlsby, N.} \emph{et~al.}
\newblock \bibinfo{title}{Parameter-efficient transfer learning for {NLP}}.
\newblock In \emph{\bibinfo{booktitle}{International Conference on Machine Learning}}, \bibinfo{pages}{2790--2799} (\bibinfo{organization}{PMLR}, \bibinfo{year}{2019}).

\bibitem{hu2021lora}
\bibinfo{author}{Hu, E.~J.} \emph{et~al.}
\newblock \bibinfo{journal}{\bibinfo{title}{{LoRA}: Low-rank adaptation of large language models}}.
\newblock {\emph{\JournalTitle{arXiv preprint arXiv:2106.09685}}}  (\bibinfo{year}{2021}).

\bibitem{qin2021learning}
\bibinfo{author}{Qin, G.} \& \bibinfo{author}{Eisner, J.}
\newblock \bibinfo{journal}{\bibinfo{title}{Learning how to ask: Querying {LMs} with mixtures of soft prompts}}.
\newblock {\emph{\JournalTitle{arXiv preprint arXiv:2104.06599}}}  (\bibinfo{year}{2021}).

\bibitem{lester2021power}
\bibinfo{author}{Lester, B.}, \bibinfo{author}{Al-Rfou, R.} \& \bibinfo{author}{Constant, N.}
\newblock \bibinfo{journal}{\bibinfo{title}{The power of scale for parameter-efficient prompt tuning}}.
\newblock {\emph{\JournalTitle{arXiv preprint arXiv:2104.08691}}}  (\bibinfo{year}{2021}).

\bibitem{li2021prefix}
\bibinfo{author}{Li, X.~L.} \& \bibinfo{author}{Liang, P.}
\newblock \bibinfo{journal}{\bibinfo{title}{Prefix-tuning: Optimizing continuous prompts for generation}}.
\newblock {\emph{\JournalTitle{arXiv preprint arXiv:2101.00190}}}  (\bibinfo{year}{2021}).

\bibitem{liu2021p}
\bibinfo{author}{Liu, X.} \emph{et~al.}
\newblock \bibinfo{journal}{\bibinfo{title}{P-tuning v2: Prompt tuning can be comparable to fine-tuning universally across scales and tasks}}.
\newblock {\emph{\JournalTitle{arXiv preprint arXiv:2110.07602}}}  (\bibinfo{year}{2021}).

\bibitem{an2022input}
\bibinfo{author}{An, S.} \emph{et~al.}
\newblock \bibinfo{journal}{\bibinfo{title}{Input-tuning: Adapting unfamiliar inputs to frozen pretrained models}}.
\newblock {\emph{\JournalTitle{arXiv preprint arXiv:2203.03131}}}  (\bibinfo{year}{2022}).

\bibitem{liu2022few}
\bibinfo{author}{Liu, H.} \emph{et~al.}
\newblock \bibinfo{journal}{\bibinfo{title}{Few-shot parameter-efficient fine-tuning is better and cheaper than in-context learning}}.
\newblock {\emph{\JournalTitle{Advances in Neural Information Processing Systems}}} \textbf{\bibinfo{volume}{35}}, \bibinfo{pages}{1950--1965} (\bibinfo{year}{2022}).

\bibitem{zhang2023fedpetuning}
\bibinfo{author}{Zhang, Z.} \emph{et~al.}
\newblock \bibinfo{title}{{FedPETuning}: When federated learning meets the parameter-efficient tuning methods of pre-trained language models}.
\newblock In \emph{\bibinfo{booktitle}{Annual Meeting of the Association of Computational Linguistics 2023}}, \bibinfo{pages}{9963--9977} (\bibinfo{organization}{Association for Computational Linguistics (ACL)}, \bibinfo{year}{2023}).

\bibitem{zhao2023fedprompt}
\bibinfo{author}{Zhao, H.}, \bibinfo{author}{Du, W.}, \bibinfo{author}{Li, F.}, \bibinfo{author}{Li, P.} \& \bibinfo{author}{Liu, G.}
\newblock \bibinfo{title}{{FedPrompt}: Communication-efficient and privacy-preserving prompt tuning in federated learning}.
\newblock In \emph{\bibinfo{booktitle}{ICASSP 2023-2023 IEEE International Conference on Acoustics, Speech and Signal Processing (ICASSP)}}, \bibinfo{pages}{1--5} (\bibinfo{organization}{IEEE}, \bibinfo{year}{2023}).

\bibitem{zhang2023towards}
\bibinfo{author}{Zhang, J.} \emph{et~al.}
\newblock \bibinfo{journal}{\bibinfo{title}{Towards building the federated gpt: Federated instruction tuning}}.
\newblock {\emph{\JournalTitle{arXiv preprint arXiv:2305.05644}}}  (\bibinfo{year}{2023}).

\bibitem{guo2023promptfl}
\bibinfo{author}{Guo, T.}, \bibinfo{author}{Guo, S.}, \bibinfo{author}{Wang, J.}, \bibinfo{author}{Tang, X.} \& \bibinfo{author}{Xu, W.}
\newblock \bibinfo{journal}{\bibinfo{title}{{PromptFL}: Let federated participants cooperatively learn prompts instead of models-federated learning in age of foundation model}}.
\newblock {\emph{\JournalTitle{IEEE Transactions on Mobile Computing}}}  (\bibinfo{year}{2023}).

\bibitem{cai2022fedadapter}
\bibinfo{author}{Cai, D.}, \bibinfo{author}{Wu, Y.}, \bibinfo{author}{Wang, S.}, \bibinfo{author}{Lin, F.~X.} \& \bibinfo{author}{Xu, M.}
\newblock \bibinfo{title}{{FedAdapter}: Efficient federated learning for modern {NLP}}.
\newblock In \emph{\bibinfo{booktitle}{ACM 29th Annual International Conference on Mobile Computing and Networking (MobiCom)}} (\bibinfo{year}{2023}).

\bibitem{smith2022using}
\bibinfo{author}{Smith, S.} \emph{et~al.}
\newblock \bibinfo{journal}{\bibinfo{title}{Using deepspeed and megatron to train megatron-turing nlg 530b, a large-scale generative language model}}.
\newblock {\emph{\JournalTitle{arXiv preprint arXiv:2201.11990}}}  (\bibinfo{year}{2022}).

\bibitem{xiao2023smoothquant}
\bibinfo{author}{Xiao, G.} \emph{et~al.}
\newblock \bibinfo{title}{{SmoothQuant}: Accurate and efficient post-training quantization for large language models}.
\newblock In \emph{\bibinfo{booktitle}{International Conference on Machine Learning}}, \bibinfo{pages}{38087--38099} (\bibinfo{organization}{PMLR}, \bibinfo{year}{2023}).

\bibitem{xiao2023offsite}
\bibinfo{author}{Xiao, G.}, \bibinfo{author}{Lin, J.} \& \bibinfo{author}{Han, S.}
\newblock \bibinfo{journal}{\bibinfo{title}{Offsite-tuning: Transfer learning without full model}}.
\newblock {\emph{\JournalTitle{arXiv preprint arXiv:2302.04870}}}  (\bibinfo{year}{2023}).

\bibitem{ding2023dc}
\bibinfo{author}{Ding, Y.} \emph{et~al.}
\newblock \bibinfo{journal}{\bibinfo{title}{{DC-CCL}: Device-cloud collaborative controlled learning for large vision models}}.
\newblock {\emph{\JournalTitle{arXiv preprint arXiv:2303.10361}}}  (\bibinfo{year}{2023}).

\bibitem{kuang2023federatedscope}
\bibinfo{author}{Kuang, W.} \emph{et~al.}
\newblock \bibinfo{journal}{\bibinfo{title}{{FederatedScope-LLM}: A comprehensive package for fine-tuning large language models in federated learning}}.
\newblock {\emph{\JournalTitle{arXiv preprint arXiv:2309.00363}}}  (\bibinfo{year}{2023}).

\bibitem{han2015deep}
\bibinfo{author}{Han, S.}, \bibinfo{author}{Mao, H.} \& \bibinfo{author}{Dally, W.~J.}
\newblock \bibinfo{journal}{\bibinfo{title}{Deep compression: Compressing deep neural networks with pruning, trained quantization and huffman coding}}.
\newblock {\emph{\JournalTitle{arXiv preprint arXiv:1510.00149}}}  (\bibinfo{year}{2015}).

\bibitem{sajjad2023effect}
\bibinfo{author}{Sajjad, H.}, \bibinfo{author}{Dalvi, F.}, \bibinfo{author}{Durrani, N.} \& \bibinfo{author}{Nakov, P.}
\newblock \bibinfo{journal}{\bibinfo{title}{On the effect of dropping layers of pre-trained transformer models}}.
\newblock {\emph{\JournalTitle{Computer Speech \& Language}}} \textbf{\bibinfo{volume}{77}}, \bibinfo{pages}{101429} (\bibinfo{year}{2023}).

\bibitem{hinton2015distilling}
\bibinfo{author}{Hinton, G.}, \bibinfo{author}{Vinyals, O.} \& \bibinfo{author}{Dean, J.}
\newblock \bibinfo{journal}{\bibinfo{title}{Distilling the knowledge in a neural network}}.
\newblock {\emph{\JournalTitle{arXiv preprint arXiv:1503.02531}}}  (\bibinfo{year}{2015}).

\bibitem{zhang2020accelerating}
\bibinfo{author}{Zhang, M.} \& \bibinfo{author}{He, Y.}
\newblock \bibinfo{journal}{\bibinfo{title}{Accelerating training of transformer-based language models with progressive layer dropping}}.
\newblock {\emph{\JournalTitle{Advances in Neural Information Processing Systems}}} \textbf{\bibinfo{volume}{33}}, \bibinfo{pages}{14011--14023} (\bibinfo{year}{2020}).

\bibitem{FedPET}
\bibinfo{author}{Zhang, Z.} \emph{et~al.}
\newblock \bibinfo{title}{{F}ed{PET}uning: When federated learning meets the parameter-efficient tuning methods of pre-trained language models}.
\newblock In \emph{\bibinfo{booktitle}{Findings of the Association for Computational Linguistics: ACL 2023}}, \bibinfo{pages}{9963--9977}, \doiprefix\url{10.18653/v1/2023.findings-acl.632} (\bibinfo{publisher}{Association for Computational Linguistics}, \bibinfo{address}{Toronto, Canada}, \bibinfo{year}{2023}).

\bibitem{FedGPT}
\bibinfo{author}{Zhang, J.} \emph{et~al.}
\newblock \bibinfo{journal}{\bibinfo{title}{Towards building the federated {GPT}: Federated instruction tuning}}.
\newblock {\emph{\JournalTitle{arXiv preprint arXiv:2305.05644}}}  (\bibinfo{year}{2023}).

\bibitem{lim2021dynamic}
\bibinfo{author}{Lim, W. Y.~B.} \emph{et~al.}
\newblock \bibinfo{journal}{\bibinfo{title}{Dynamic edge association and resource allocation in self-organizing hierarchical federated learning networks}}.
\newblock {\emph{\JournalTitle{IEEE Journal on Selected Areas in Communications}}} \textbf{\bibinfo{volume}{39}}, \bibinfo{pages}{3640--3653} (\bibinfo{year}{2021}).

\bibitem{rician}
\bibinfo{author}{Xiao, C.}, \bibinfo{author}{Zheng, Y.~R.} \& \bibinfo{author}{Beaulieu, N.~C.}
\newblock \bibinfo{title}{Statistical simulation models for rayleigh and rician fading}.
\newblock In \emph{\bibinfo{booktitle}{IEEE International Conference on Communications, 2003. ICC'03.}}, vol.~\bibinfo{volume}{5}, \bibinfo{pages}{3524--3529} (\bibinfo{organization}{IEEE}, \bibinfo{year}{2003}).

\bibitem{PPO}
\bibinfo{author}{Schulman, J.}, \bibinfo{author}{Wolski, F.}, \bibinfo{author}{Dhariwal, P.}, \bibinfo{author}{Radford, A.} \& \bibinfo{author}{Klimov, O.}
\newblock \bibinfo{journal}{\bibinfo{title}{Proximal policy optimization algorithms}}.
\newblock {\emph{\JournalTitle{arXiv preprint arXiv:1707.06347}}}  (\bibinfo{year}{2017}).

\bibitem{kahn1951estimation}
\bibinfo{author}{Kahn, H.} \& \bibinfo{author}{Harris, T.~E.}
\newblock \bibinfo{journal}{\bibinfo{title}{Estimation of particle transmission by random sampling}}.
\newblock {\emph{\JournalTitle{National Bureau of Standards Applied Mathematics Series}}} \textbf{\bibinfo{volume}{12}}, \bibinfo{pages}{27--30} (\bibinfo{year}{1951}).

\bibitem{tavakoli2018action}
\bibinfo{author}{Tavakoli, A.}, \bibinfo{author}{Pardo, F.} \& \bibinfo{author}{Kormushev, P.}
\newblock \bibinfo{title}{Action branching architectures for deep reinforcement learning}.
\newblock In \emph{\bibinfo{booktitle}{Proceedings of the aaai conference on artificial intelligence}}, vol.~\bibinfo{volume}{32} (\bibinfo{year}{2018}).

\bibitem{schulman2015high}
\bibinfo{author}{Schulman, J.}, \bibinfo{author}{Moritz, P.}, \bibinfo{author}{Levine, S.}, \bibinfo{author}{Jordan, M.} \& \bibinfo{author}{Abbeel, P.}
\newblock \bibinfo{journal}{\bibinfo{title}{High-dimensional continuous control using generalized advantage estimation}}.
\newblock {\emph{\JournalTitle{arXiv preprint arXiv:1506.02438}}}  (\bibinfo{year}{2015}).

\bibitem{erceg1999empirically}
\bibinfo{author}{Erceg, V.} \emph{et~al.}
\newblock \bibinfo{journal}{\bibinfo{title}{An empirically based path loss model for wireless channels in suburban environments}}.
\newblock {\emph{\JournalTitle{IEEE Journal on selected areas in communications}}} \textbf{\bibinfo{volume}{17}}, \bibinfo{pages}{1205--1211} (\bibinfo{year}{1999}).

\bibitem{kingma2014adam}
\bibinfo{author}{Kingma, D.~P.} \& \bibinfo{author}{Ba, J.}
\newblock \bibinfo{journal}{\bibinfo{title}{Adam: A method for stochastic optimization}}.
\newblock {\emph{\JournalTitle{arXiv preprint arXiv:1412.6980}}}  (\bibinfo{year}{2014}).

\bibitem{lowe2017multi}
\bibinfo{author}{Lowe, R.} \emph{et~al.}
\newblock \bibinfo{journal}{\bibinfo{title}{Multi-agent actor-critic for mixed cooperative-competitive environments}}.
\newblock {\emph{\JournalTitle{Advances in neural information processing systems}}} \textbf{\bibinfo{volume}{30}} (\bibinfo{year}{2017}).

\end{thebibliography}
\if
\section*{Acknowledgements}

This research is supported in part by Nanyang Technological
University (NTU), the NTU-Wallenberg AI, Autonomous Systems and Software
Program (WASP) Joint Project; NTU Startup Grant; the Sin-
gapore Ministry of Education Academic Research Fund under
Grant Tier 1 RG97/20, Grant Tier 1 RG24/20 and Grant Tier
2 MOE2019-T2-1-176.

\section*{Author contributions statement}

T.J.C, WH.Y, Y.L, and J.Z contributed equally and wrote the main manuscript text and software of programming. 

\section*{Competing interests statement}

The authors declare no competing interests.
\fi
\section*{Legends}
\textbf{Figure 1. }Intersection of Federated learning (FL), Parameter-Efficient Fine-Tuning (PEFT), and Emulator-Assisted Tuning (\mbox{\texttt{EAT}}). Here we illustrate the intersection of FL, PEFT, and (\mbox{\texttt{EAT}}). The main contribution of our current paper is to introduce Federated Parameter-Efficient Emulator-Assisted Tuning (\mbox{\texttt{FedPEAT}}), as a convergence of \mbox{\texttt{EAT}}, PEFT, and FL, while \mbox{\texttt{EAT}} and Parameter-Efficient Emulator-Assisted Tuning (\mbox{\texttt{PEAT}}) are also terms coined by our paper.\\

\noindent\textbf{Figure 2. } \mbox{\texttt{FedPEAT}} with Adaptive control overview. This figure shows how the Adaptive control orchestrator makes decisions on important parameters, such as device selection, emulator compression parameter, transmission bandwidth and power to facilitate the \mbox{\texttt{FedPEAT}} process.\\

\noindent\textbf{Figure 3. }Emulator-Assisted Tuning generalized to three cases. Figure illustrates how the neural network structures at the server and local devices differ in each case. Case 1 represents our proposed \mbox{\texttt{FedPEAT}} framework. Case 2 represents the integration of Federated Learning and PEFT. Case 3 represents a traditional Federated Learning scenario.\\

\noindent\textbf{Figure 4. }Our proposed SABPPO algorithm and architecture. Figure illustrates the underlying actor and critic architecture, their interaction with the environment and model update process.\\

\noindent\textbf{Figure 5. }Comparison between FL and \mbox{\texttt{FedPEAT}}, and Comparison between Adaptive control algorithms. Figure \ref{fig:fl_delay}, \ref{fig:fl_emuxchange}, \ref{fig:fl_perplexity} illustrates the performance difference between FL and \mbox{\texttt{FedPEAT}} with regards to delay, emulator exchange count, and perplexity, respectively. Figure~\ref{fig:traintime} illustrates the time taken for model training for each adaptive control algorithm. Figure \ref{fig:fed_delay}, \ref{fig:fed_emuxchange}, \ref{fig:fed_perplexity}, \ref{fig:fed_reward} illustrate the performance of each adaptive control algorithm in across the training process, in terms of log(delay), emulator exchange count, perplexity, and reward.\\

\noindent\textbf{Algorithm 1. }\mbox{\texttt{FedPEAT}} with adaptive control mechanism.\\

\noindent\textbf{Algorithm 2. }SABPPO adaptive control algorithm.

\end{document}